\pdfoutput=1
\PassOptionsToPackage{table}{xcolor}

\documentclass[sigconf]{acmart}

\AtBeginDocument{}

\usepackage{amssymb}

\usepackage{colortbl}

\definecolor{green}{RGB}{168,208,141}
\definecolor{blue}{RGB}{144,170,219}
\definecolor{orange}{RGB}{244,177,131}

\usepackage{booktabs}   
\usepackage{multirow}
\usepackage{graphicx}   

\renewcommand\footnotetextcopyrightpermission[1]{}

\begin{document}

\title{Mitigating Query Selection Bias in Referring Video Object Segmentation}

\author{Dingwei Zhang}

\affiliation{%
\institution{Nanjing University of Science and Technology, Nanjing, China}
  \country{}
}

\author{Dong Zhang}
\affiliation{%
  \institution{The Hong Kong University of Science and Technology, Hong Kong, China}
    \country{}
  }

\author{Jinhui Tang}

\affiliation{%
\institution{Nanjing Forestry University,\\ Nanjing, China}
\country{}
  }

\renewcommand{\shortauthors}{Dingwei Zhang, Dong Zhang, \& Jinhui Tang.}

\begin{abstract}
Recently, query-based methods have achieved remarkable performance in Referring Video Object Segmentation (RVOS) by using textual static object queries to drive cross-modal alignment.
However, these static queries are easily misled by distractors with similar appearance or motion, resulting in \emph{query selection bias}.
To address this issue, we propose Triple Query Former (TQF), which factorizes the referring query into three specialized components: an appearance query for static attributes, an intra-frame interaction query for spatial relations, and an inter-frame motion query for temporal association.  Instead of relying solely on textual embeddings, our queries are dynamically constructed by integrating both linguistic cues and visual guidance. Furthermore, we introduce two motion-aware aggregation modules that enhance object token representations: Intra-frame Interaction Aggregation incorporates position-aware interactions among objects within a single frame, while Inter-frame Motion Aggregation leverages trajectory-guided alignment across frames to ensure temporal coherence. Extensive experiments on multiple RVOS benchmarks demonstrate the advantages of TQF and the effectiveness of our structured query design and motion-aware aggregation modules. 
\end{abstract}



\maketitle

\section{Introduction}
\label{intro}
Referring Video Object Segmentation (RVOS) is a core visual grounding task involving natural language understanding and visual perception, with the goal of segmenting referred objects across the frames. 
Compared to Semi-supervised VOS~\cite{chen2023boosting}, the key challenge in RVOS lies not only in handling variations in object shape, occlusion, and complex background distractions but also in accurately identifing the referred object based on language descriptions. 

Recently, query-based methods, represented by ReferFormer~\cite{wu2022language}, have achieved remarkable advancements in RVOS benchmarks\cite{mao2016generation,ding2023mevis,yu2016modeling,Seo2020}. ReferFormer\cite{wu2022language} and its variants\cite{he2024decoupling,han2023html,yuan2024losh,yan2024referred,li2023robust,Wu_2023_ICCV,mei2024slvp,luo2024soc,miao2023spectrum,tang2023temporal,hu2024temporal,li2024univs}, built on DETR\cite{carion2020end}, leverage linguistic descriptions as object query conditions to establish a more efficient cross-modal interaction. 

\begin{figure}[!t]
\centering
\includegraphics[width=\linewidth]{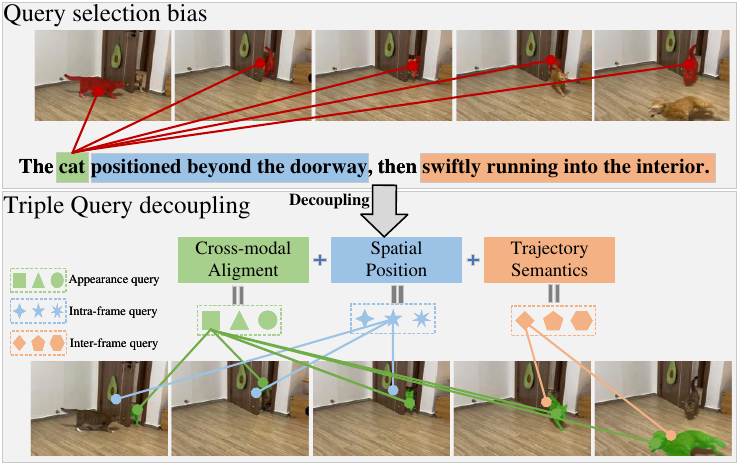}
\caption{
\textbf{Illustration of query selection bias and our Triple Query Former (TQF).}
Static text-initialized queries (top) mislead the model when encountering distractors with similar appearance or motion.
Our TQF (bottom) decouples the sentence into three specialized queries:
\textcolor{green}{appearance} (object identity), \textcolor{blue}{intra-frame} (spatial relation), and \textcolor{orange}{inter-frame} (temporal behavior),
each query generation and selection via cross-modal attention for more precise grounding.
}
\label{fig:query_bias_tqf}
\end{figure}

However, above methods often suffer from \textit{query selection bias}. As illustrated in Figure~\ref{fig:query_bias_tqf}, when background objects share similar appearances to the target, these methods tend to overly rely on appearance-based textual cues, leading to segmentation errors. This issue stems from the fact that traditional approaches typically treat the entire referring expression as a set of static object queries, which is uniformly reused across all frames. Nevertheless, referring expressions often encode diverse semantic components, including object attributes, spatial relations, and temporal interactions. Relying on a fixed query across time restricts the model’s capacity to adaptively leverage these different linguistic cues.

Although  DsHmp~\cite{he2024decoupling} attempts to decouple static and motion cues for query generation and thereby reduce the over-reliance on appearance, the resulting queries are still initialized solely  from textual features. This text-only initialization lacks grounding in the visual space and fails to capture semantic and motion cues from visual spatiotemporal embedding.  When  multiple objects  exhibit comparable motion behaviors, this simple decoupling shifts the bias from appearance to motion. The dominance of motion semantics in such cases may lead to misidentification, as the model overlooks spatial or contextual differences. In fact, rigid or misaligned decoupling may even lead to degraded performance, yielding results inferior to those obtained with undivided sentence-level queries.

To address the aforementioned limitations, we propose \textbf{Triple Query Former (TQF)} decoupling the referring query into three complementary components. The \textbf{appearance query} is constructed in the visual feature space under the guidance of static textual attributes, enabling the frame-level segmenter to extract object-oriented tokens without relying solely on text embeddings. The \textbf{intra-frame query} models relational interactions among objects in the same frame using action-related language cues. The \textbf{inter-frame  query} integrates trajectory embeddings and temporal action semantics to capture long-range motion patterns across frames. To further exploit the structured query design, we introduce two motion-aware refinement modules: \textbf{Intra-frame Interaction Aggregation (IIA)} and \textbf{Inter-frame Motion Aggregation (IMA)}. IIA  enhances object token representations by constructing pairwise directed relational features based on spatial geometry and attention cues, and selectively aggregating those that are semantically aligned with intra-frame interaction descriptions.
IMA refines object token trajectories by combining local temporal attention and semantic-guided inter-frame alignment, ensuring consistent identity and semantics across time. 

Extensive experiments on challenging RVOS benchmarks, including A2D-Sentences~\cite{gavrilyuk2018actor}, JHMDB-Sentences~\cite{gavrilyuk2018actor}, Ref-DAVIS17~\cite{khoreva2019video}, Ref-YouTube-VOS~\cite{Seo2020}, and MeViS~\cite{ding2023mevis}, demonstrate the superiority of our proposed TQF. Under both Video-Swin-T and Video-Swin-B backbones, TQF consistently achieves state-of-the-art performance across all datasets. Compared to previous best methods, it brings significant improvements in terms of region accuracy, contour quality, and temporal consistency. Notably, TQF surpasses DsHmp~\cite{he2024decoupling} and LoSh~\cite{yuan2024losh} by over \textbf{+3\%} and \textbf{+2\%} on Ref-DAVIS17 and Ref-YouTube-VOS respectively, and achieves up to \textbf{+21.8\%} gain over ReferFormer~\cite{wu2022language} on MeViS.

The main contributions of this work can be summarized as follows:
\begin{itemize}
    \item We propose Triple Query Former (TQF), a unified framework that decouples the referring query into appearance, intra-frame, and inter-frame components, enabling fine-grained modeling of static attributes, spatial relations, and temporal motion in RVOS.
    \item We introduce two motion-aware refinement modules—Intra-frame Interaction Aggregation (IIA) and Inter-frame Motion Aggregation (IMA)—which enhance the  discriminability and consistency of object token representations by leveraging relational priors and cross-modal temporal motion alignment.
    \item Extensive experiments on multiple RVOS benchmarks demonstrate the effectiveness of our approach. 
\end{itemize}

\section{Related work}
\label{related work}

\subsection{One-stage Methods} One-stage RVOS methods predict masks directly through multimodal interaction between visual and textual inputs. Wang et al. \cite{wang2019asymmetric} introduce cross-modal attention for actor-action segmentation. Liu et al. \cite{liu2021cross} enhance segmentation with spatial-temporal reasoning and text-guided feature exchange. Ding et al. \cite{ding2022language} propose language-bridging for efficient spatiotemporal encoding. Chen et al. \cite{chen2022multi} employ multi-attention networks in compressed video for efficiency. Wu et al. \cite{wu2022multi} focus on dynamic semantic alignment of long-term objects. YOFO \cite{li2022you} utilizes meta-learning for targeted language cues. MTTR \cite{zhao2022modeling} integrates appearance, flow, and linguistic features using transformers. However, these methods perform early-stage pixel-level fusion, limiting their use of object-level cues and cross-frame information.
\subsection{Two-stage Methods} 
Two-stage RVOS methods segment object masks first and then ground masks to textual descriptions \cite{liang2021rethinking,liang2021clawcranenet}. While this pipeline effectively enhances semantic understanding between text and visual information, it introduces several challenges. Firstly, the complexity of the model increases significantly because the segmentation and grounding models must be trained and optimized separately, which demands greater computational resources and time. Secondly, inference becomes more cumbersome, as visual and textual information must be processed independently, potentially hindering real-time performance. Moreover, the independent modules in this two-stage pipeline are susceptible to information loss or noise, which can limit the potential for performance gains in end-to-end optimization.


\subsection{Query-based Methods}

To address the limitations of both one-stage and two-stage methods, query-based approaches have emerged as a promising alternative. Inspired by recent semantic segmentation methods\cite{zhang2020feature, chen2024dual,chen2024transformer,zhang2020causal,chen2024learning,wang2023coupling,wang2024boosting}, query-based methods introduce object queries that enable more flexible handling of cross-frame objects and actions, while fostering efficient interactions between visual and linguistic modalities\cite{dai2025accctr,dai2025noisectrl,wang2025emcontrol}.  Adam et al. \cite{botach2022end} propose an end-to-end multimodal transformer that frames RVOS as a sequence prediction problem. ReferFormer \cite{wu2022language} pioneers the use of language queries to directly focus on target regions, generating segmentation masks via dynamic convolution kernels and enabling object tracking. R²-VOS \cite{li2023robust} introduces cycle consistency to address object-text misalignment in video segmentation. TempCD \cite{tang2023temporal} designs a temporal collect-and-distribute mechanism for efficient spatiotemporal reasoning by exchanging local and global information between reference labels and object queries. SgMg \cite{miao2023spectrum} mitigates feature drift with a spectrum-guided multi-granularity segmentation paradigm, enabling global intra-frame interaction in the spectral domain. UniRef++ \cite{Wu_2023_ICCV} unifies RVOS, RIS, and VOS via a multi-path fusion approach for instance-level segmentation. HTML \cite{han2023html} integrates language and visual features across temporal scales to extract core object semantics. SOC \cite{luo2024soc} introduces Semantic-assisted Object Clustering, enhancing temporal modeling and cross-modal alignment for better video-level visual-linguistic integration. LoSH \cite{yuan2024losh} combines long- and short-text visual modeling with temporal consistency loss to reduce bias toward action- or relation-based expressions. However, most query-based methods segment frames independently using static textual queries, limiting their ability to capture spatiotemporal consistency. DsHmp \cite{he2024decoupling} addresses this by decoupling static and dynamic textual cues and hierarchically perceiving motion over time. 
Nevertheless, its static query design hinders adaptability to dynamic object changes, and its computational complexity limits efficiency and scalability in long videos and multi-target scenarios.

  \begin{figure*}[!t]         
  \centering         
  \includegraphics[width=\textwidth]{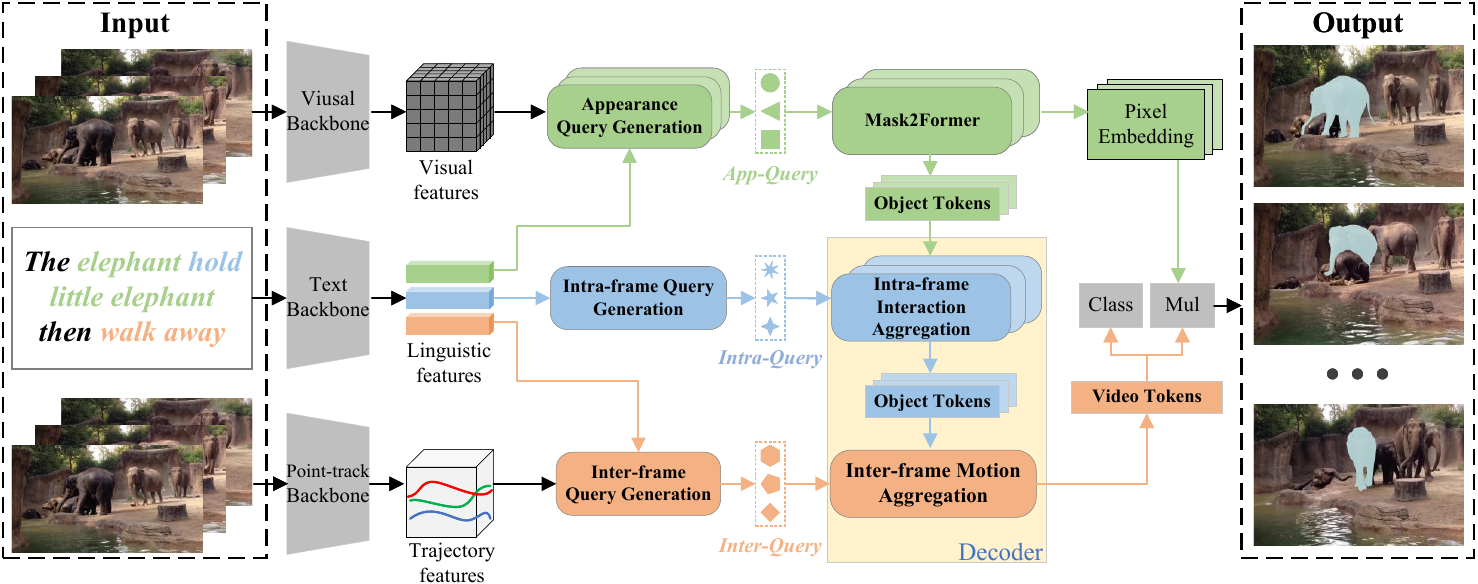}             
  \caption{Overview of our proposed Triple Query Former (TQF). TQF decouples the linguistic description into three parts: \textcolor{green}{appearance attributes (green)}, \textcolor{blue}{interaction semantics (blue)} and \textcolor{orange}{temporal motion semantics (orange)}. Appearance Query consists of static features with visual features, and generates object tokens and pixel representations via Mask2Former. In the decoder, Intra-frame Interaction Aggregation module first takes the interactive motion features as Intra-frame Query to establish correspondence between in-frame object tokens. Subsequently, Inter-frame Motion Aggregation module captures cross-frame motion associations using Inter-frame Query derived from trajectory and independent motion features, producing video tokens for final mask prediction. Best viewed in color.}        
  \label{fig:framework}      
  \end{figure*} 
 
\section{Method}
\label{method}

The overview of our proposed method as shown in Figure \ref{fig:framework}. Given a video clip $\mathcal{V} = \{v_t\}^T$, $ v_t\in \mathbb{R}^{3 \times H \times W}$ and its corresponding textual description $\mathcal{D}=\{d_w\}^W$ with $W$ words, the objective is to produce referring object segmentation mask $\mathcal{S}=\{s_t\}^T$ of each frame. To achieve this, we propose a Triple Query Former (TQF), which is built on top of VITA\cite{heo2022vita} and ReferFormer \cite{wu2022language}. 

\subsection{Backbone}\label{3.1}

\noindent\textbf{Visual Encoder.} 
 For a given video clip with $T$ frames, we utilize a pre-trained visual encoder to extract spatiotemporal representations, denoted as $ \mathbf{F}_{v}^{i} \in \mathbb{R}^{T \times C_i \times H_i \times W_i}$. Here, $i \in \{1, 2, 3, 4\}$ corresponds to different feature levels, $H_i$and $W_i$ denote the spatial dimensions, and $C_i$ is the channel dimension. 
 


\noindent\textbf{Text Encoder.}  Given a textual description containing \(W\) words, we first feed the entire sentence into a pre-trained RoBERTa\cite{liu2019roberta} to obtain contextualized word embeddings $\mathbf{F}_l \in \mathbb{R}^{W \times D}$,  where \(W\) is the number of words in the sentence, and \(D\) is the hidden dimension of the textual features. This representation captures rich global context and token-level semantics across the whole sentence.
 To extract fine-grained semantic components, we utilize spaCy\cite{Honnibal_spaCy_Industrial-strength_Natural_2020} to identify three types of linguistic phrases from the sentence: \emph{static descriptions} (e.g., object category, color, shape), \emph{interactive motions} (e.g., inter-object relationships), and \emph{independent motions} (e.g., temporal dynamics of a single object). Based on the token spans returned by spaCy, we generate corresponding binary masks \(\mathcal{M}^s, \mathcal{M}^r, \mathcal{M}^e \in \{0,1\}^W\), and extract the following subset features from \(\mathbf{F}_l\): 
\begin{equation} 
\mathbf{F}_l^s = \mathbf{F}_l[\mathcal{M}^s], \quad \mathbf{F}_l^r = \mathbf{F}_l[\mathcal{M}^r], \quad \mathbf{F}_l^e = \mathbf{F}_l[\mathcal{M}^e],
\end{equation} 
where \(\mathbf{F}_l^s \in \mathbb{R}^{W_s \times D}\), \(\mathbf{F}_l^r \in \mathbb{R}^{W_r \times D}\), and \(\mathbf{F}_l^e \in \mathbb{R}^{W_e \times D}\) correspond to the static, relational, and temporal subcomponents of the textual input, respectively. These decomposed features are utilized by downstream modules to guide intra- and inter-frame query generation.

\noindent\textbf{Point Tracker.} In the first frame, we select \(N_E\) pixels based on the cross-modal similarity between visual and static textual features. Subsequently, these pixels are tracked frame by frame using a pre-trained Point Tracker, which extracts their corresponding trajectory coordinates \(\mathcal{T} = \{(x_t, y_t)\}_{t=0}^{T-1}\) as well as trajectory embeddings \(\mathcal{Z}\in \mathbb{R}^{N_E \times T \times C}\), where \(C\) indicates the dimensionality of the visual features.

\subsection{Triple Query Generation}\label{3.2}
Query-based methods typically initialize the linguistic description as video object queries, which are repeated in each frames. While this way provides a straightforward mapping between textual and visual content, it often struggles to effectively capture the complex motions in the video sequence. To address this issues, we introduce three distinct types of query: Appearance Query, Intra-frame Query, and Inter-frame Query. Each query type is designed to target a specific aspect of the visual-textual alignment process, ensuring comprehensive feature representation. 


\noindent\textbf{Appearance Query Generation.} 
In contrast to DsHmp\cite{he2024decoupling}, which initializes appearance queries from textual static attributes,  our method derives core representations of the object's static attributes directly from the visual features of the current frame. This visually driven approach is inherently more compatible with single-modal segmentation models like Mask2Former. Although generating appearance queries through cross-modal attention across the entire feature hierarchy is straightforward, it incurs significant computational overhead at low-level features, where semantic information is limited, reducing the effectiveness of static attribute extraction. To address this, we propose a Top-K-based hierarchical text-to-image cross-modal attention mechanism. Our method first aligns high-level features cross-modally to identify the most relevant regions. Subsequent fine-grained computations are then confined to these regions at mid- and low-level features. This hierarchical approach not only reduces computational cost but also improves the efficiency of static attribute capture.

In particular, we first compute the preliminary alignment \(\mathbf{H}^4\) by applying cross attention (CA) between the static textual attribute features \(\mathbf{F}_l^s\) and the highest-level visual feature \(\mathbf{F}_v^4\) as follows:
\begin{equation}
    \mathbf{H}^4 =\mathbf{F}_l^s + \text{MLP}\bigl( \text{CA}({W}^{Q}\mathbf{F}_l^s,W^{K}\mathbf{F}_v^i, {W}^{V}\mathbf{F}_v^i) \bigr), 
\end{equation}
where ${W}^{Q}$, ${W}^{K}$ and ${W}^{V}$ denote linear projection matrices.

For the lower-level visual(i.e. $i \in \{1,2,3\}$) features, the hierarchical alignment process is defined as:
\begin{equation}
    \mathbf{H}^i =\mathbf{H}^{i+1} +\text{MLP} \bigl( \text{CA}(\mathbf{H}^{i+1}, {W}^{K}\mathbf{F}_v^i[\mathcal{I}^{i+1}], {W}^{V}\mathbf{F}_v^i[\mathcal{I}^{i+1}]) \bigr), 
\end{equation}
where $\mathcal{I}^{i+1}$ represents the Top-K most relevant position indices obtained from the previous alignment \(\mathbf{H}^{i+1}\), which guide the selection of key regions in the visual feature space. This ensures that attention at the $i$-th level focuses only on the most important regions identified from the higher level, progressively narrowing down the computational scope while preserving the regions critical for extracting static attributes.

Finally, we compute cross-modal attention between the lowest-level alignment \(\mathbf{H}^1\) and a set of learnable initial queries \(\mathcal{Q}^{\text{Init}}_{\text{App}} \), generating \(N_A\) appearance queries $\mathcal{Q}_{\text{App}}\in \mathbb{R}^{N_A \times C}$:

\begin{equation}
\mathcal{Q}_{\text{App}} = \mathcal{Q}^{\text{Init}}_{\text{App}}+\text{MLP}\bigl(\text{CA}(\mathcal{Q}^{\text{Init}}_{\text{App}},W^{K}\mathbf{H}^1, {W}^{V}\mathbf{H}^1)\bigr).
\end{equation}

These appearance queries encode a compact and semantically enriched representation of the object’s static attributes, which have been progressively refined through the hierarchical Top-K-guided cross-modal attention mechanism.

\noindent\textbf{Intra-frame Query Generation.}
To effectively capture the interactions between different objects within a single frame, we propose the generation of specialized intra-frame queries $\mathcal{Q}_{\text{Intra}} \in \mathbb{R}^{N_I\times C}$. These queries are explicitly designed to encode spatial relationships between object tokens by jointly modeling interaction-relevant motion priors and local contextual semantics. As spatial relationships within a single frame tend to be stable and exhibit minimal temporal variability, these intra-frame queries can be efficiently reused across multiple frames. 

Specifically, we construct the intra-frame queries $\mathcal{Q}_{\text{Intra}}$ by integrating two complementary sources of information. On the one hand, we utilize a set of learnable initial query vectors $\mathcal{Q}^{\text{Init}}_{\text{Intra}} \in \mathbb{R}^{N_I \times C}$, which act as high-level interaction slots to capture potential object-object relationships. On the other hand, we extract interaction-relevant textual features $\mathbf{F}_l^{r}$ from referring expressions that describe spatial or functional relationships between objects (e.g., "hold"). A cross-attention mechanism is applied to inject semantic information from $\mathbf{F}_l^{r}$ into the initial queries, yielding enhanced interaction-aware representations.
 To ensure these queries remain grounded in the overall scene semantics, the attention output is further concatenated with the full textual representation $\mathbf{F}_l$. The resulting vector is passed through a multilayer perceptron (MLP) and added to the initial query via a residual connection, yielding the final intra-frame queries:

\begin{equation}     
\mathcal{Q}_{\text{Intra}} = \mathcal{Q}^{\text{Init}}_{\text{Intra}}    + \text{MLP}\bigl(\text{Concat}[     \text{CA}(\mathcal{Q}^{\text{Init}}_{\text{Intra}}, W^{K}\mathbf{F}_l^{r}, W^{V}\mathbf{F}_l^{r}),      \mathbf{F}_l]\bigr).
\end{equation} 

By construction, $\mathcal{Q}_{\text{Intra}}$ encodes both explicit interaction cues and context-aware semantics, making them ideal for driving spatial relation modeling across all frames in a consistent and efficient manner.

\noindent\textbf{Inter-frame Query Generation.}
Unlike intra-frame interactions that rely primarily on spatial relations, cross-frame object modeling places greater emphasis on maintaining the temporal continuity and semantic consistency of tokens across time. Achieving stable associations of the same semantic tokens across different frames requires the effective fusion of motion trajectory information and visual semantic features, enabling a unified representation of the object's temporal evolution and semantic identity. To this end, we propose a  inter-frame query, which explicitly models the correspondence between object motion trajectories and action-level semantics, thereby generating a global representation with semantic selection capability.

Considering the temporal redundancy inherent in the trajectory embeddings $\mathcal{Z}$ obtained from the Point Tracker, we adopt an attention pooling to perform compact aggregation and extract the most discriminative motion features. Specifically, we first introduce a set of learnable initial query vectors and perform cross-attention with the encoded textual features $\mathbf{F}_l^{e}$, which represent independently described motion semantics. This yields a set of semantically enriched query vectors $\mathcal{A} \in \mathbb{R}^{N_E \times C}$ as follows:
\begin{equation}
    \mathcal{A} = \mathcal{Q}_{\text{Inter}}^{\text{Init}} + \mathrm{MLP}\left( \mathrm{CA}\left( \mathcal{Q}_{\text{Inter}}^{\text{Init}},\ W^K \mathbf{F}_l^{e},\ W^V \mathbf{F}_l^{e} \right) \right).
\end{equation}

To enhance the spatiotemporal sensitivity of the attention pooling process, we introduce explicit positional encodings for both the query vectors $\mathcal{A}$ and the trajectory embeddings $\mathcal{Z}$, denoted as $\mathbf{P}_\mathcal{A}$ and $\mathbf{P}_\mathcal{Z}$, respectively: 

\begin{equation} 
\mathbf{P}_\mathcal{A} = \text{Embed} + \mathrm{MLP}(\mathcal{A}), \quad  \mathbf{P}_\mathcal{Z} = \mathrm{SineCos}(\mathcal{T}) + \mathrm{MLP}(\mathcal{Z}),
\end{equation}
where $\text{Embed}$ is a learnable positional embedding that provides query-wise identity encoding, and $\mathrm{SineCos}(\mathcal{T})$ represents a temporal sinusoidal encoding based on the trajectory coordinates $\mathcal{T}$.

Finally, we perform attention pooling on the trajectory embeddings using the semantically enhanced query vectors. This results in a compact representation $\mathcal{Q}_{\text{Inter}} \in \mathbb{R}^{N_E \times C}$ that effectively fuses visual motion patterns with independent motion semantics:
\begin{equation}
\mathcal{Q}_{\text{Inter}} = \mathrm{Softmax}\left( (\mathcal{A} + P_{\mathcal{A}})(\mathcal{Z} + P_{\mathcal{Z}})^\top \right)\, \mathcal{Z}.
\end{equation}

\subsection{Spatiotemporal Token Aggregation}\label{3.3}
\begin{figure}[!t]     \centering     \includegraphics[width=\linewidth]{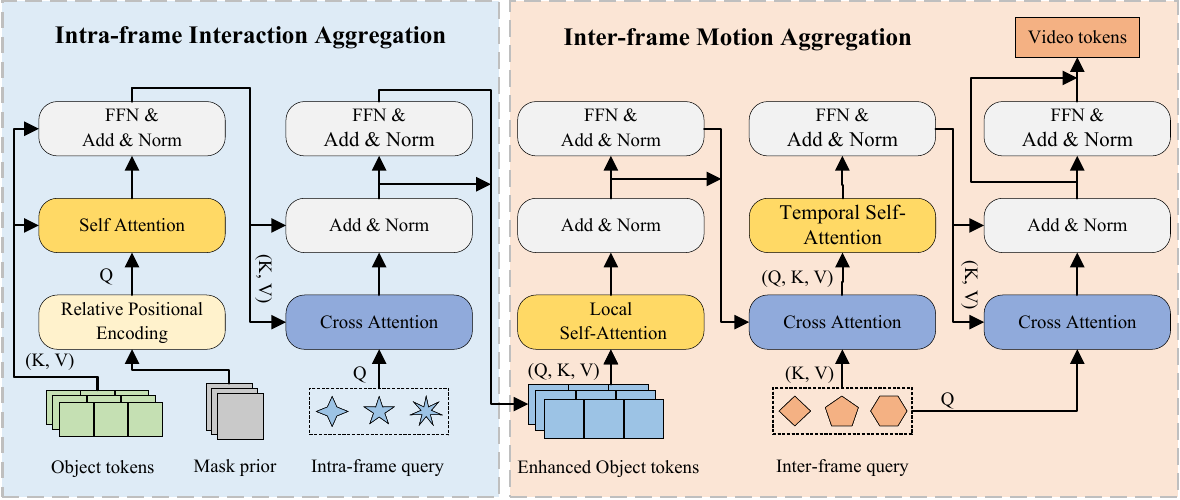}     \caption{The pipeline of Intra-frame Interaction Aggregation and Inter-frame Motion Aggregation modules}     \label{fig:motion_decoder}  \end{figure} 
Following previous works~\cite{gao2023decoupling, heo2022vita}, we adopt a pre-trained Mask2Former \cite{cheng2022masked}, which takes appearance queries $\mathcal{Q}_{\text{App}}$ and visual features $\mathbf{F}_v$ as input to generate a set of potential object tokens $\mathcal{O} \in \mathbb{R}^{N_A \times T \times C}$, their corresponding coarse masks, and fine-grained pixel embeddings $\mathcal{M} \in \mathbb{R}^{T \times C_1 \times H \times W}$.

\noindent\textbf{Intra-frame Interaction Aggregation.} 
Although Mask2Former\cite{cheng2022masked} enables effective extraction of object-oriented tokens, its inherent design as a single-frame segmenter limits its ability to capture motion-related relationships between objects. In Referring Video Object Segmentation (RVOS), the target object described in the text often exhibits explicit interactions or motion relationships with other objects, such as “a person riding a bicycle” or “a person playing basketball.” These relational cues are crucial for accurately understanding the video content. Without effectively modeling the alignment between inter-object motion relations in both textual and visual spaces, the model may struggle to distinguish foreground from background, leading to ambiguous semantic grounding. To address this limitation, we introduce the Intra-frame Interaction Aggregation module, which explicitly models the motion semantics among object tokens within a single frame using text-guided intra-frame queries.

For the initial object tokens, we construct a total of $N(N-1)$ directed intra-frame relational features $\mathcal{R}_{i \to j} \in \mathbb{R}^{C}$, which encoding the motion-related interactions between each tokens. These  relational features incorporate both spatial geometric relationships and motion interaction cues. Specifically, we first leverage the mask priors from Mask2Former  and explicitly encode the relative spatial position between tokens in the same frame:

\begin{equation}
    \text{RelPos}_{i \to j}= \left[ \log\left( \frac{|x_i - x_j|}{w_i} + 1 \right),\; \log\left( \frac{|y_i - y_j|}{h_i} + 1 \right),\; \mathrm{IoU}_{ij} \right] ,
\end{equation}
where $|x_i - x_j|$ and $|y_i - y_j|$ represent the absolute differences between the centroids of object masks. These differences are normalized by the width $w_i$ and height $h_i$ to ensure scale invariance, and $\mathrm{IoU}_{ij}$ denotes the Intersection-over-Union between the two object masks, capturing their spatial overlap.

To better align the spatial geometric relations with intra-frame interaction semantics, we apply a sine-cosine positional encoding  to project the relative position vector into a high-dimensional space. This is followed by a multi-layer perceptron (MLP) that performs nonlinear transformation on the encoded features. To ensure non-negativity of the resulting representation and to mitigate potential gradient vanishing issues in subsequent attention computations, we further apply a ReLU activation function.  The resulting enhanced spatial relation feature between tokens $\mathcal{O}_i$ and $\mathcal{O}_j$ is denoted as $\phi_{i \to j}^{\text{RelPos}}$:
\begin{equation}
\phi_{i \to j}^{\text{RelPos}} = \mathrm{ReLU}\left( \mathrm{MLP}\left( \mathrm{SineCos}\left( \mathrm{RelPos}_{i \to j} \right) \right) \right).
\end{equation}

Subsequently,  we  incorporate $\phi_{i \to j}^{\text{Rel-Pos}}$ as an explicit spatial prior into the self-attention computation among intra-frame object tokens. The attention from token $\mathcal{O}_i$ to token $\mathcal{O}_j$ is computed as:  
\begin{equation}
\mathcal{R}_{i \to j} = \mathrm{Softmax}\left( \phi_{i \to j}^{\text{Rel-Pos}} + \mathcal{O}_i \cdot \mathcal{O}_j \right)\mathcal{O}_j.
\end{equation} 

In this way, each object token can adaptively focus on other tokens that are closely related to their own spatial geometric, leading to more effective feature aggregation and more precise modeling of motion interactions. 

Finally,  we measure the semantic alignment between each relational feature $\mathcal{R}_{i \to j}$ and the intra-frame query to select the object interactions that are most consistent with the described action in the visual-semantic space. This mechanism not only suppresses interference from background or irrelevant object relations, but also highlights the key action participants and their interaction patterns within the frame. The selected relation features are then weighted and fused with the original object tokens to obtain enhanced representations that incorporate inter-object interactions:
\begin{equation}
    \mathcal{O}'_i = \mathcal{O}_i + \text{MLP}(\text{CA}(\mathcal{Q}_\text{Intra}, W^K\mathcal{R}_{i \to j} , W^V\mathcal{R}_{i\to j} )),
\end{equation}
where the matrices $W^K$  and $W^V$  used here and in subsequent equations are independently learned parameters and thus are not identical.

\noindent\textbf{Inter-frame Motion Aggregation.} 
To construct a cross-frame sequence of object tokens, a common approach is to use the Hungarian algorithm for establishing inter-frame correspondences. However, due to significant appearance variations across frames, static matching based solely on visual similarity often leads to inconsistent spatiotemporal representations.

Consequently, we introduce the Inter-frame Motion Aggregation module, which combines global semantic awareness with local temporal modeling. As shown in the figure, we first apply localized inter-frame self-attention to capture temporal context across adjacent frames and enhance representation consistency: 
\begin{equation} 
\tilde{\mathcal{O}}_i[t] = \textstyle \sum_{k \in \mathcal{N}(t)} \mathrm{Softmax}\left( W^Q \mathcal{O}'_i[t] \cdot W^K \mathcal{O}'_j[k] \right) W^V \mathcal{O}'_j[k],
\end{equation} 
where $\mathcal{O}'_j[k]$ represents the initial object token matched to $\tilde{\mathcal{O}}_i[t]$ at frame $k$, obtained via the Hungarian algorithm.  $\mathcal{N}(t)$ denotes a temporal neighborhood centered at frame $t$ , typically with a window size of 3 (i.e., ${t-1,t,t+1}$).

Subsequently, we leverage cross-modal attention to guide the object tokens toward alignment with long-range semantic priors. This process compensates for appearance-level gaps that arise when visual cues are unreliable but semantic continuity remains, thereby encouraging the object representations to converge into a unified, action-aware semantic space. Furthermore, we apply a temporal self-attention mechanism over tokens corresponding to the same object  across frames to enhance semantic consistency across time. The overall process can be formally expressed as: 
\begin{equation} 
\widehat{\mathcal{O}} = \mathrm{TempAttn} \left( \mathrm{CA} \left( \tilde{\mathcal{O}},\, W^K\mathcal{Q}_{\text{Inter}},\, W^V\mathcal{Q}_{\text{Inter}} \right) \right).
\end{equation} 
This unified refinement guides the tokens converge toward a unified global motion-aware semantic space, even when visual consistency is not explicitly observable.

Finally, we repurpose the inter-frame query from a passive semantic guide to an active semantic selector. Each inter-frame query is designed to represent a potential motion semantic or behavioral pattern, and should be capable of selectively attending to object tokens that are highly relevant to its own semantics—while ignoring irrelevant or noisy tokens. By doing so, it produces a set of video-level representations that captures the most salient motion semantics:
\begin{equation}
   \mathcal{V} = \mathrm{FFN} \left( \mathrm{CA} \left( \mathcal{Q}_{\text{Inter}}, W^K\widehat{\mathcal{O}}, W^V\widehat{\mathcal{O}}  \right) \right).
\end{equation}

\subsection{Prediction Head and Loss function}\label{3.4}
We adopt a prediction head to generate the final segmentation results based on the learned video-level tokens and pixel-level embeddings. Specifically, the refined video tokens $\mathcal{V} \in \mathbb{R}^{N \times C}$ are first projected into class and mask coefficients through two separate linear layers. To determine the final output, we rank all predicted mask tracklets according to their classification confidence scores and retain the top-ranked ones. This selection strategy effectively suppresses background noise and ensures that only the most semantically relevant object masks are preserved in the final prediction.

Inspired by~\cite{heo2022vita}, we adopt a multi-branch training objective that jointly supervises both the per-frame and video-level predictions. The overall loss consists of three components: a frame-level loss \(\mathcal{L}_{\text{f}}\), computed from the output of the frame-wise decoder and supervised by the corresponding ground-truth masks; a video-level loss \(\mathcal{L}_{\text{v}}\), calculated from the video token outputs and their temporally aligned ground truth; and a semantic consistency loss \(\mathcal{L}_{\text{sim}}\), which encourages temporal coherence by enforcing consistent predictions for the same object across frames.
To explicitly promote smooth temporal transitions of token representations, the semantic consistency loss is formulated as: \begin{equation} 
 \mathcal{L}_{\text{con}} = \frac{1}{T} \sum_{t=0}^{T-1} \left(1 - \cos\left(\widehat{\mathcal{O}}_i^t,\, \widehat{\mathcal{O}}_i^{t+1}\right)\right).
 \end{equation} 
 This loss enforces alignment in the feature space for temporally adjacent object tokens of the same object, even in the presence of appearance variations.
 
 The final training objective integrates all components in a weighted form:
 \begin{equation} \mathcal{L}_{\text{total}} = \lambda_{\text{v}} \cdot \mathcal{L}_{\text{v}} + \lambda_{\text{f}} \cdot \mathcal{L}_{\text{f}} + \lambda_{\text{con}} \cdot \mathcal{L}_{\text{con}},
 \end{equation}
 where \(\lambda_{\text{v}}, \lambda_{\text{f}}, \lambda_{\text{con}}\) are hyperparameters controlling the contributions of each component.

\begin{table*}[htbp]
\centering
\caption{Comparison with the query-based methods on  various benchmarks. \textbf{Bold} and \underline{underlined} results indicate the best and second-best performance, respectively.}
\label{tab1}
\resizebox{\textwidth}{!}{
\begin{tabular}{lcccc ccc ccc ccc}
\toprule
\multirow{2}{*}{Method} & \multirow{2}{*}{Reference} 
& \multicolumn{3}{c}{A2D-Sentences} 
& \multicolumn{3}{c}{JHMDB-Sentences} 
& \multicolumn{3}{c}{Ref-DAVIS17} 
& \multicolumn{3}{c}{Ref-YouTube-VOS} \\
\cmidrule(lr){3-5} \cmidrule(lr){6-8} \cmidrule(lr){9-11} \cmidrule(lr){12-14}
& & mAP & oIoU & mIoU & mAP & oIoU & mIoU 
& $\mathcal{J}$ & $\mathcal{F}$ & $\mathcal{J}\&\mathcal{F}$ 
& $\mathcal{J}$ & $\mathcal{F}$ & $\mathcal{J}\&\mathcal{F}$ \\
\midrule
\multicolumn{14}{c}{\textit{Video-Swin-T / Swin-T}} \\
\midrule

ReferFormer\cite{wu2022language} & CVPR'22 & 52.8 & 77.6 & 69.6 & 42.2 & 71.9 & 71.0 & 56.5 & 62.7 & 59.6 & 58.0 & 60.9 & 59.4 \\
         HTML\cite{han2023html} & ICCV'23 & 53.4 & 77.6 & 69.2 & 42.7 & - &-  & - &-  &-  & 59.5 & 63.0 & 61.2 \\
         SgMg\cite{miao2023spectrum} & ICCV'23 & 56.1 & 78.0 & 70.4 & 44.4 & 72.8 & 71.7 & 59.0 & 64.8 & 61.9 & 60.4 & 63.5 & 62.0 \\
    R$^2$-VOS \cite{li2023robust}& ICCV'23 &  - & - & - &-  & - & - &  -& - & - & 59.6 & 63.1 & 61.3 \\
       TempCD \cite{tang2023temporal}& ICCV'23 &  -&  -&  -& - & - &-  & 59.3 & 65.0 & 62.2 & 60.5 & 64.0 & 62.3 \\
          SOC \cite{luo2024soc}& NeurIPS'23 & 54.8 & 78.3 & 70.6 & 42.7 & 72.7 & 71.6 & 60.2 & 66.7 & 63.5 & 61.1 & 63.7 & 62.4 \\
         LoSh \cite{yuan2024losh} & CVPR'24 & \underline{57.6} & \underline{79.3} & \underline{71.6} & - & - & - & 60.1 & 65.7 & 62.9 & \underline{62.0} & \underline{65.4} & \underline{63.7} \\
        DsHmp \cite{he2024decoupling}& CVPR'24 & 57.2 & 79.0 & 71.3 & \underline{44.9} & \underline{73.1} & \underline{72.1} & \underline{60.8} & \underline{67.2} & \underline{64.0} & 61.8 & \underline{65.4} & 63.6 \\
Grounded-SAM2 \cite{ren2024groundedsamassemblingopenworld}& arXiv'24 & 47.5 & 59.0 & 62.5 & 38.9 & 70.8 & 70.5 & 57.4 & 62.6 & 60.9 & 51.8 & 57.0 & 54.4 \\

\rowcolor{gray!20}
\textbf{TQF (ours)} & - & \textbf{59.2} & \textbf{80.4} & \textbf{71.9} & \textbf{45.9} & \textbf{74.0} & \textbf{73.2} & \textbf{63.4} & \textbf{70.8} & \textbf{67.1} & \textbf{63.9} & \textbf{67.7} & \textbf{65.8} \\
\midrule
\multicolumn{14}{c}{\textit{Video-Swin-B / Swin-B}} \\
\midrule

ReferFormer \cite{wu2022language}& CVPR'22 & 55.0 & 78.6 & 70.3 & 43.7 & 73.0 & 71.8 & 58.1 & 64.1 & 61.1 & 61.3 & 64.6 & 62.9 \\
HTML \cite{han2023html}& ICCV'23 & 56.7 & 79.5 & 71.2 & 44.2 & - & - & 59.2 & 65.1 & 62.1 & 61.5 & 65.2 & 64.4 \\
SgMg \cite{miao2023spectrum}& ICCV'23 & 58.5 & 79.9 & 72.0 & 45.0 & 73.7 & 72.5 & 60.6 & 66.0 & 63.3 & 63.9 & 67.4 & 66.4 \\
TempCD \cite{tang2023temporal}& ICCV'23 & - & - & - & - & - & - & 61.6 & 67.6 & 64.6 & 63.6 & 68.0 & 65.8 \\
SOC \cite{luo2024soc}& NeurIPS'23 & 57.3 & 80.7 & 72.5 & 44.6 & 73.6 & 72.3 & 61.0 & 67.4 & 64.2 & 64.1 & 67.4 & 66.0 \\
LoSh \cite{yuan2024losh}& CVPR'24 & \underline{59.9} & \underline{81.2} & \underline{73.1} & - & - & - & 61.8 & 66.8 & 64.3 & \underline{65.4} & 69.0 & \underline{67.2} \\
DsHmp \cite{he2024decoupling}& CVPR'24 & 59.8 & 81.1 & 72.9 & \underline{45.8} & \underline{73.9} & \underline{72.9} & 62.1 & \underline{69.1} & \underline{65.6} & 65.0 & \underline{69.1} & 67.1 \\
Grounded-SAM2 \cite{ren2024groundedsamassemblingopenworld}& arXiv'24 & 54.7 & 67.8 & 68.5 & 42.5 & 72.1 & 72.1 & \underline{62.6} & \underline{69.7} & \underline{66.2} & 62.5 & 67.0 & 64.8 \\

\rowcolor{gray!20}
\textbf{TQF (ours)} & - & \textbf{61.3} & \textbf{82.4} & \textbf{73.3} & \textbf{46.7} & \textbf{74.5} & \textbf{73.7} & \textbf{64.2} & \textbf{73.0} & \textbf{68.6} & \textbf{67.7} & \textbf{70.5} & \textbf{69.1} \\
\bottomrule
\end{tabular}
}
\end{table*}

\section{Experiments}
\subsection{Implementation Details}



We use Video Swin Transformer~\cite{liu2021videoswintransformer} as visual backbone, pre-trained RoBERTa~\cite{liu2019roberta} as text encoder, and CoTracker3~\cite{karaev2024cotracker3} for trajectory extraction. For Ref-YouTube-VOS and A2D-Sentences, we follow a two-stage training scheme: pretraining on RefCOCO/+/g\cite{yu2016modeling,mao2016generation} for 300K iterations, followed by finetuning on the target set for 50K iterations. The trained models are directly evaluated on Ref-DAVIS17 and JHMDB-Sentences without further finetuning. For MeViS, we adopt single-stage training from scratch. 
All models are optimized using AdamW with a learning rate of 5e-5. Each clip contains 8 randomly sampled frames obtained via sliding windows. Frames are resized to have a short side of 360 and a long side capped at 640. Data augmentations include random flip, resize, crop, and photometric distortion. 
We set the number of appearance queries, intra-frame queries, and inter-frame queries to \(N_A = 16\), \(N_I = 8\), and \(N_E = 8\), respectively. In the appearance query generation, we set K as $8$ for Top-K-based hierarchical text-to-image cross-modal attention. For loss weighting, we use fixed coefficients: \(\lambda_{\text{v}} = 1.0\), \(\lambda_{\text{f}} = 1.0\), and \(\lambda_{\text{con}} = 0.5\), corresponding to video-level supervision, frame-level accuracy, and temporal consistency.

\subsection{Quantitative results}

\begin{table}[htbp]
\centering
\caption{Comparison on MeViS validation set. We report two variants: TQF\textit{-T} and TQF\textit{-B}, which adopt Video-Swin-Tiny and Video-Swin-Base backbones, respectively.}
\label{tab2}
\begin{tabular}{lcccc}
\toprule
Methods & Reference & $\mathcal{J}$ & $\mathcal{F}$ & $\mathcal{J}\&\mathcal{F}$ \\
\midrule
URVOS\cite{Seo2020} & ECCV'20 & 25.7 & 29.9 & 27.8 \\
LBDT\cite{ding2022language} & CVPR'22 & 27.8 & 30.8 & 29.3 \\
MTTR\cite{zhao2022modeling} & CVPR'22 & 28.8 & 31.2 & 30.0 \\
ReferFormer\cite{wu2022language} & CVPR'22 & 29.8 & 32.2 & 31.0 \\
VLT+TC\cite{ding2022vlt} & TPAMI'22 & 33.6 & 37.3 & 35.5 \\
LMPM\cite{ding2023mevis} & ICCV'23 & 34.2 & 40.2 & 37.2 \\
DsHmp\cite{he2024decoupling} & CVPR'24 & \underline{43.0} & \underline{49.8} & \underline{46.4} \\
Grounded-SAM2\cite{ren2024groundedsamassemblingopenworld} & arXiv'24 & 34.5 & 47.4 & 40.5 \\
\rowcolor{gray!20}
TQF\textit{-T} (ours) & - & \textbf{46.8} & \textbf{53.3} & \textbf{50.1} \\
\rowcolor{gray!20} 
TQF\textit{-B} (ours) & - & \textbf{49.4} & \textbf{56.1} & \textbf{52.8} \\
\bottomrule
\end{tabular}
\end{table}

\begin{table}[htbp]
\centering
\caption{Ablation study of TQF on Ref-YouTube-VOS. IIA: Intra-frame Interaction Aggregation; IMA: Inter-frame Motion Aggregation;  PE: Positional Encodings; RPE: Relative Positional Encodings.}
\label{tab3}
\resizebox{\linewidth}{!}{
\begin{tabular}{lccc}
\toprule
\textbf{Setting} & $\mathcal{J}$ & $\mathcal{F}$ & $\mathcal{J}\&\mathcal{F}$ \\
\midrule
\multicolumn{4}{c}{\textit{Modules Ablation}} \\
\midrule
w/o \textsc{IIA} & 55.9 \textcolor{red}{(-8.0)} & 60.5 \textcolor{red}{(-7.2)} & 58.2 \textcolor{red}{(-7.6)} \\
w/o \textsc{IMA} & 54.3 \textcolor{red}{(-9.6)} & 59.5 \textcolor{red}{(-8.2)} & 56.9 \textcolor{red}{(-8.9)} \\
w/o \textsc{IIA} \& \textsc{IMA} & 42.6 \textcolor{red}{(-21.3)} & 51.4 \textcolor{red}{(-16.3)} & 49.0 \textcolor{red}{(-16.8)} \\
IIA w/o RPE & 61.9 \textcolor{red}{(-2.0)} & 65.5 \textcolor{red}{(-2.2)} & 63.7 \textcolor{red}{(-2.1)} \\
\midrule
\multicolumn{4}{c}{\textit{Query Ablation}} \\
\midrule
$N_A$ w/o cross-modal fusion & 60.1 \textcolor{red}{(-3.8)} & 64.5 \textcolor{red}{(-3.2)} & 62.3 \textcolor{red}{(-3.5)} \\
$N_I$ w/o sentence fusion & 63.0 \textcolor{red}{(-0.9)} & 66.0 \textcolor{red}{(-1.7)} & 64.5 \textcolor{red}{(-1.3)} \\
$N_E$ w/o trajectory fusion & 58.8 \textcolor{red}{(-5.1)} & 62.0 \textcolor{red}{(-5.7)} & 60.4 \textcolor{red}{(-5.4)} \\
$N_E$ w/o PE & 62.1 \textcolor{red}{(-1.8)} & 66.3 \textcolor{red}{(-1.4)} & 64.2 \textcolor{red}{(-1.6)} \\

\midrule
\rowcolor{gray!20}
TQF (Full Model) & \textbf{63.9} & \textbf{67.7} & \textbf{65.8} \\
\bottomrule
\end{tabular}
}
\end{table}

\begin{figure*}[!h]
\centering
\setlength{\tabcolsep}{1pt} 
\renewcommand{\arraystretch}{1.6} 

\begin{tabular}{c@{\hskip 2pt}c@{\hskip 2pt}c@{\hskip 2pt}c@{\hskip 2pt}c@{\hskip 2pt}c@{\hskip 2pt}c@{\hskip 2pt}c} \makebox[0.8cm][c]{\rotatebox{90}{ \small Baseline\cite{wu2022language}}} &
\includegraphics[width=0.12\linewidth]{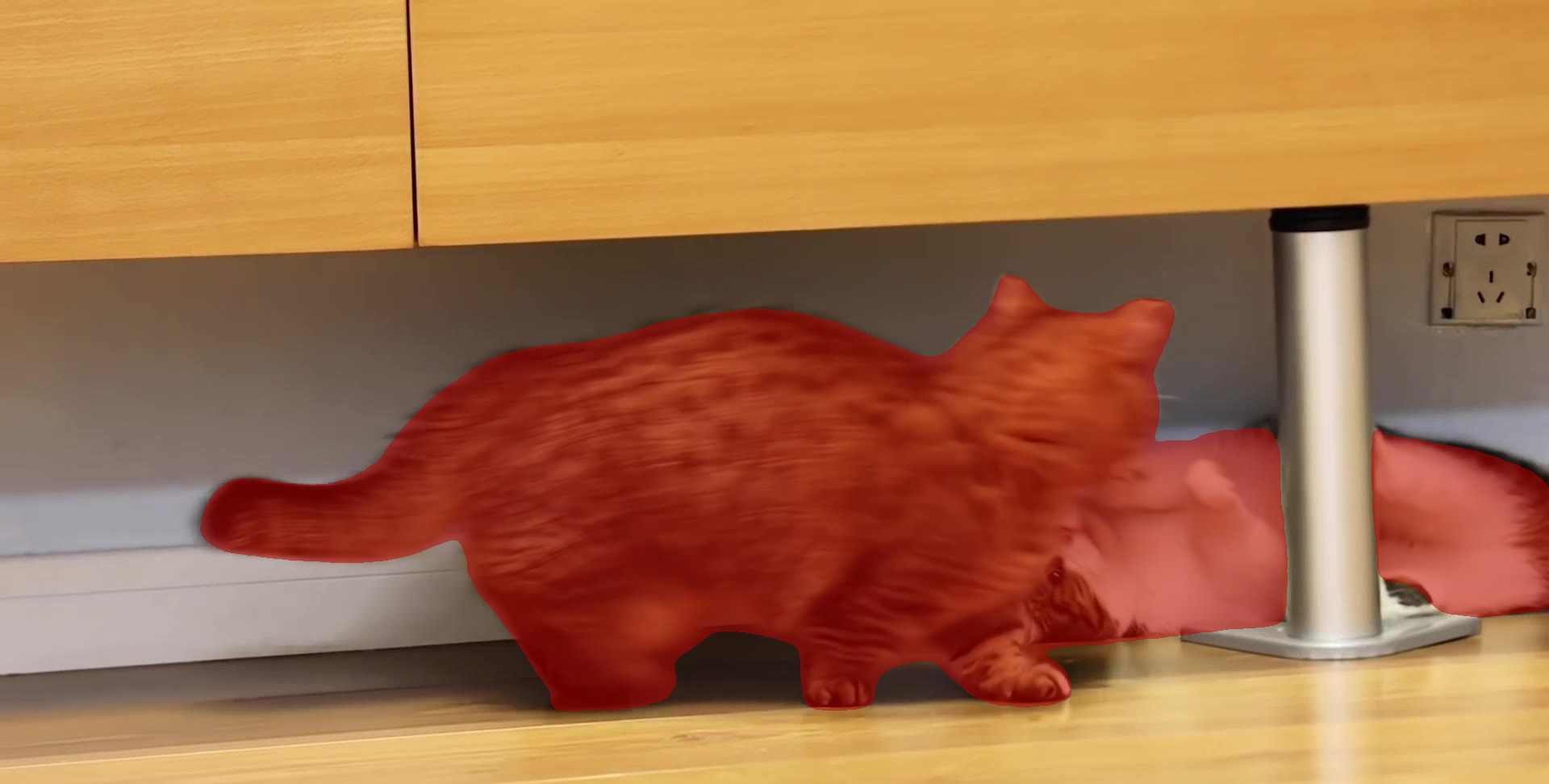} & 
\includegraphics[width=0.12\linewidth]{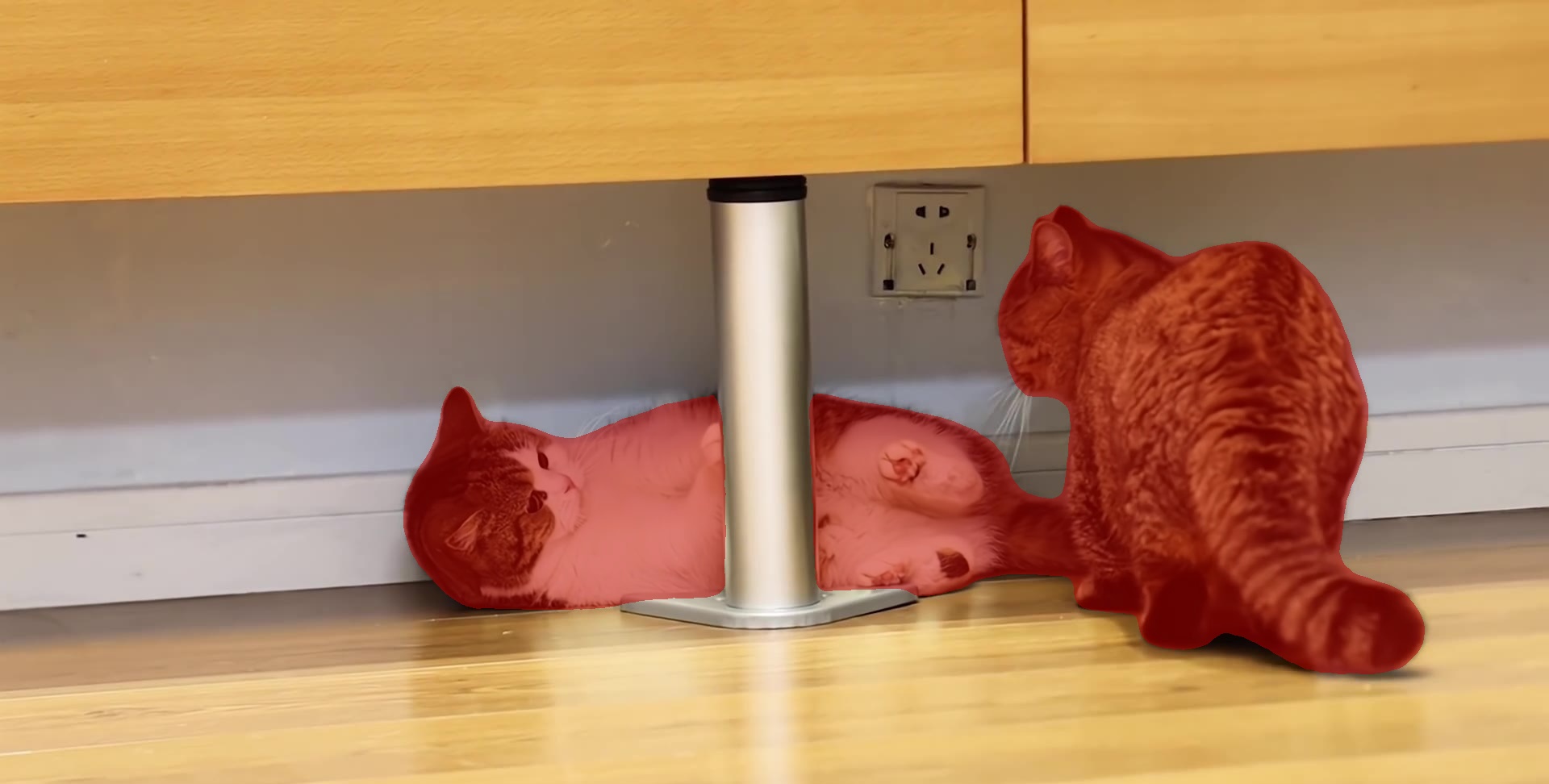} &
\includegraphics[width=0.12\linewidth]{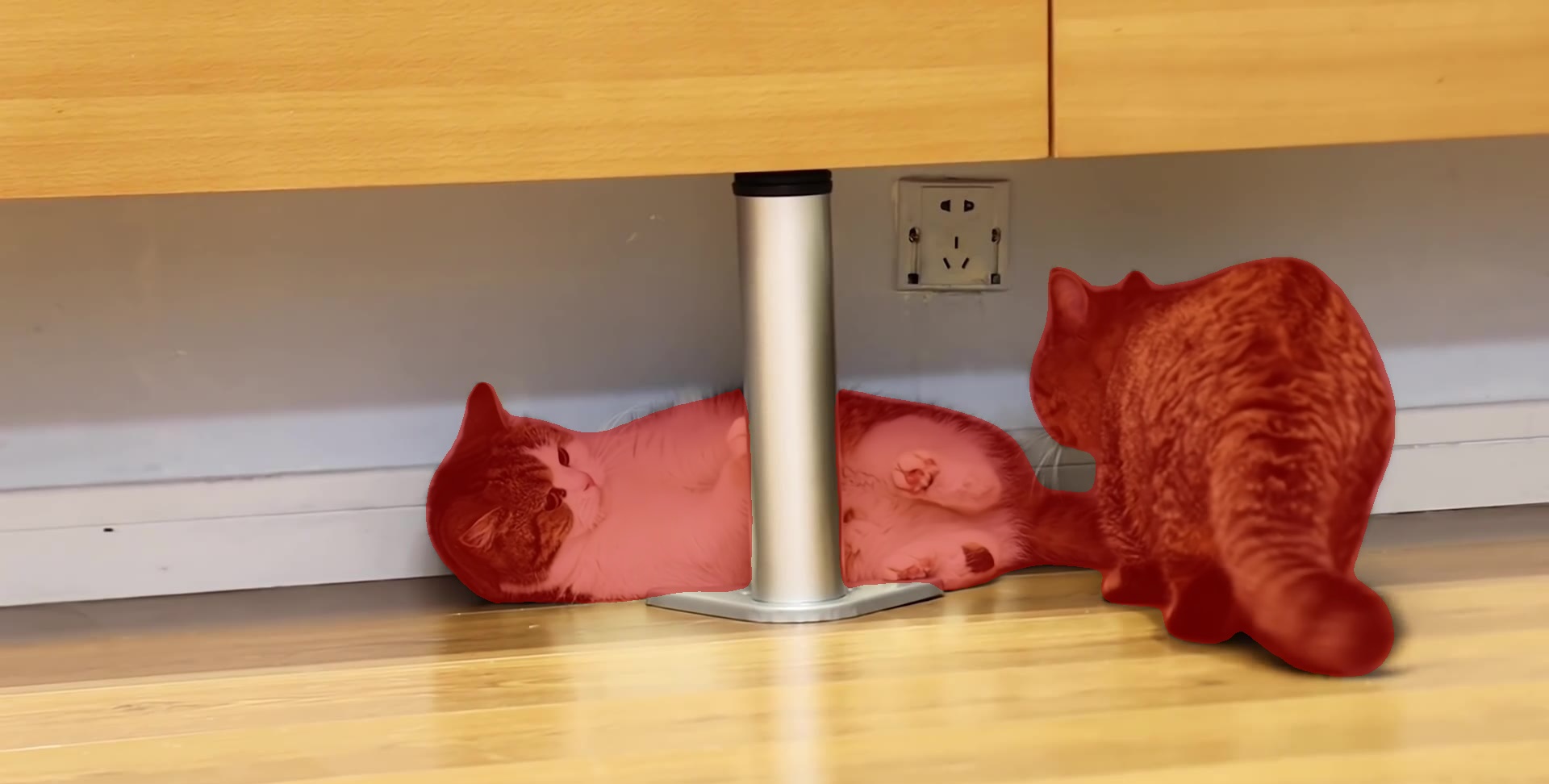} & 
\includegraphics[width=0.12\linewidth]{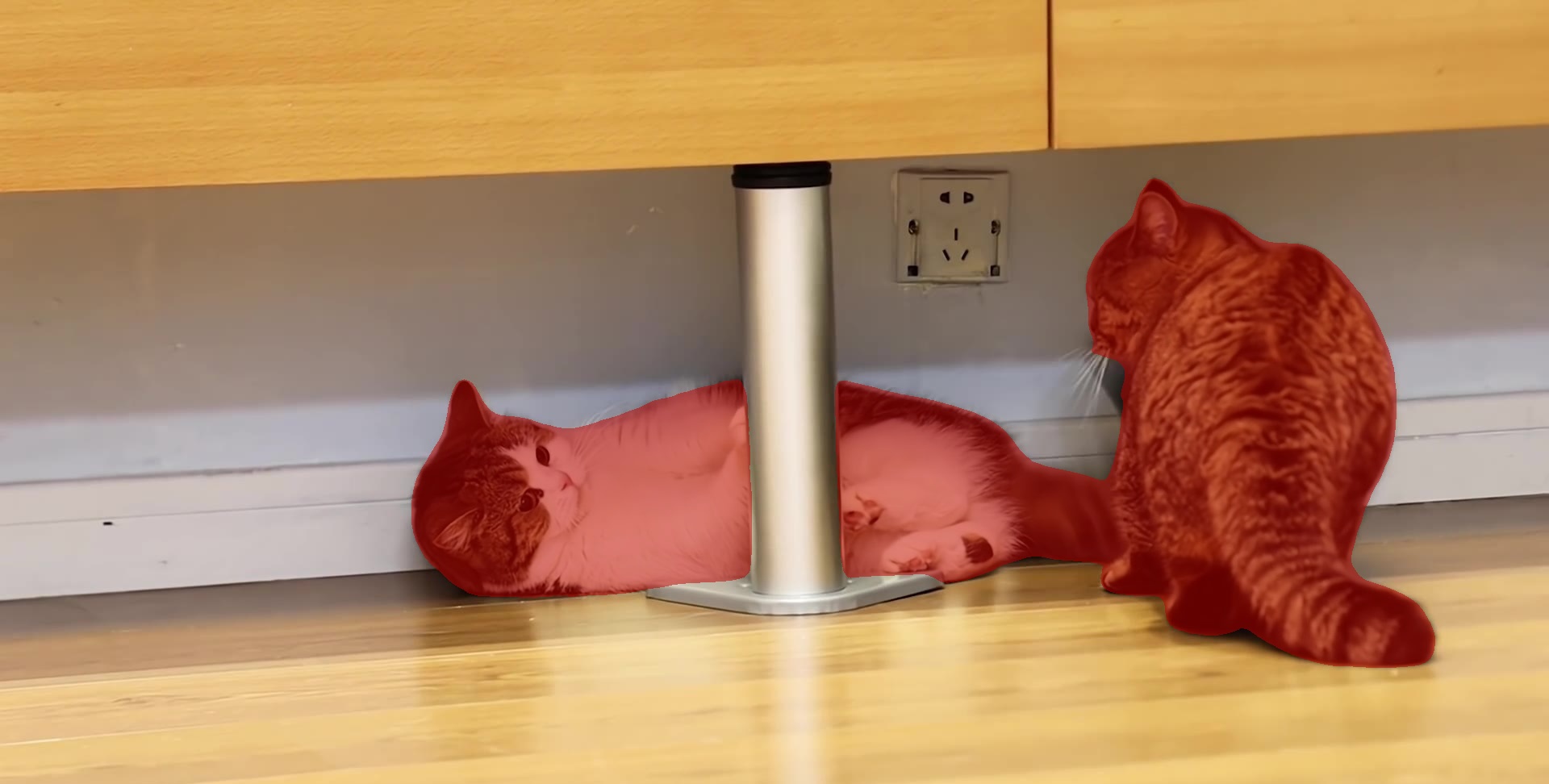} & 
\includegraphics[width=0.12\linewidth]{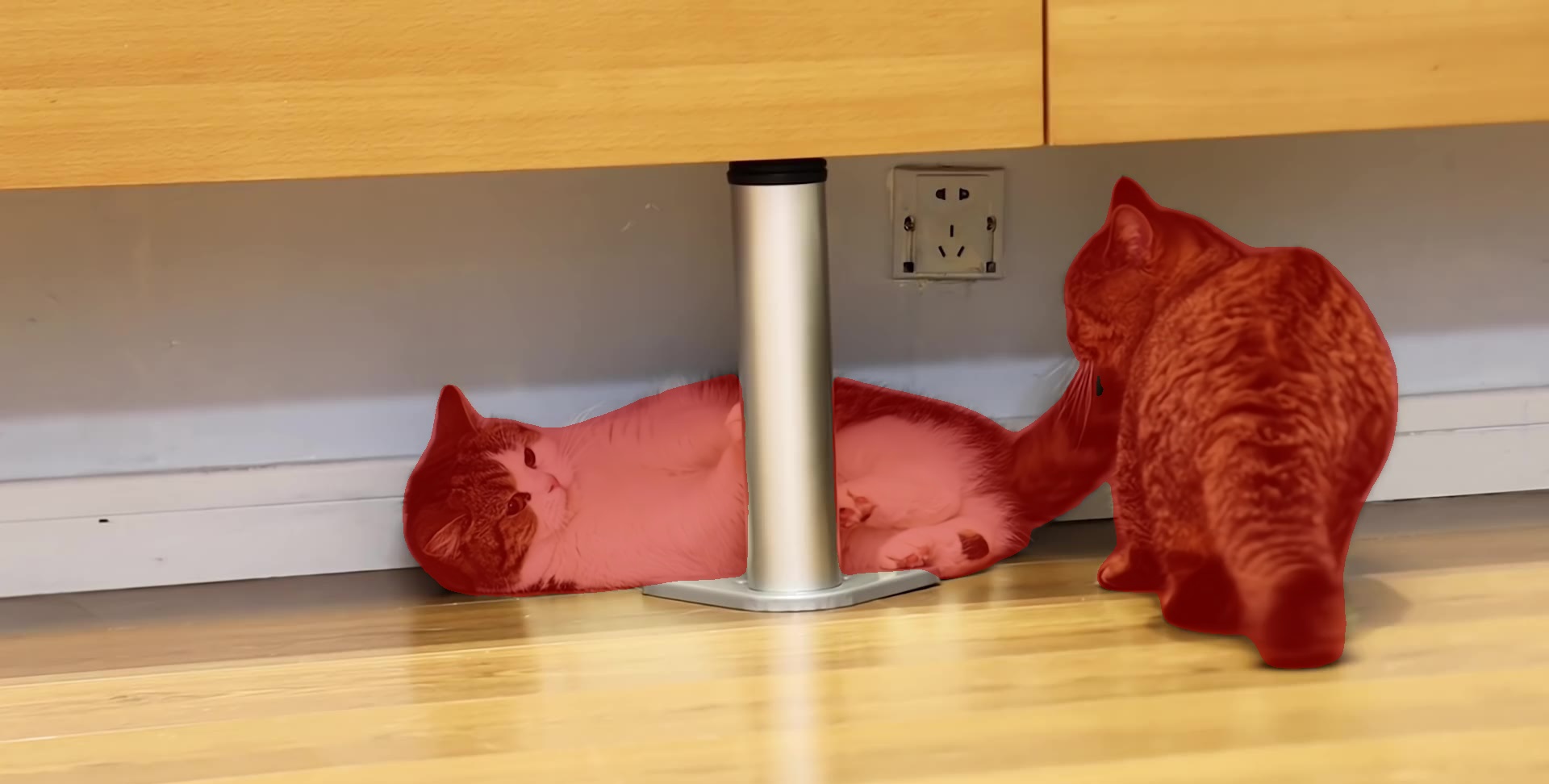} & 
\includegraphics[width=0.12\linewidth]{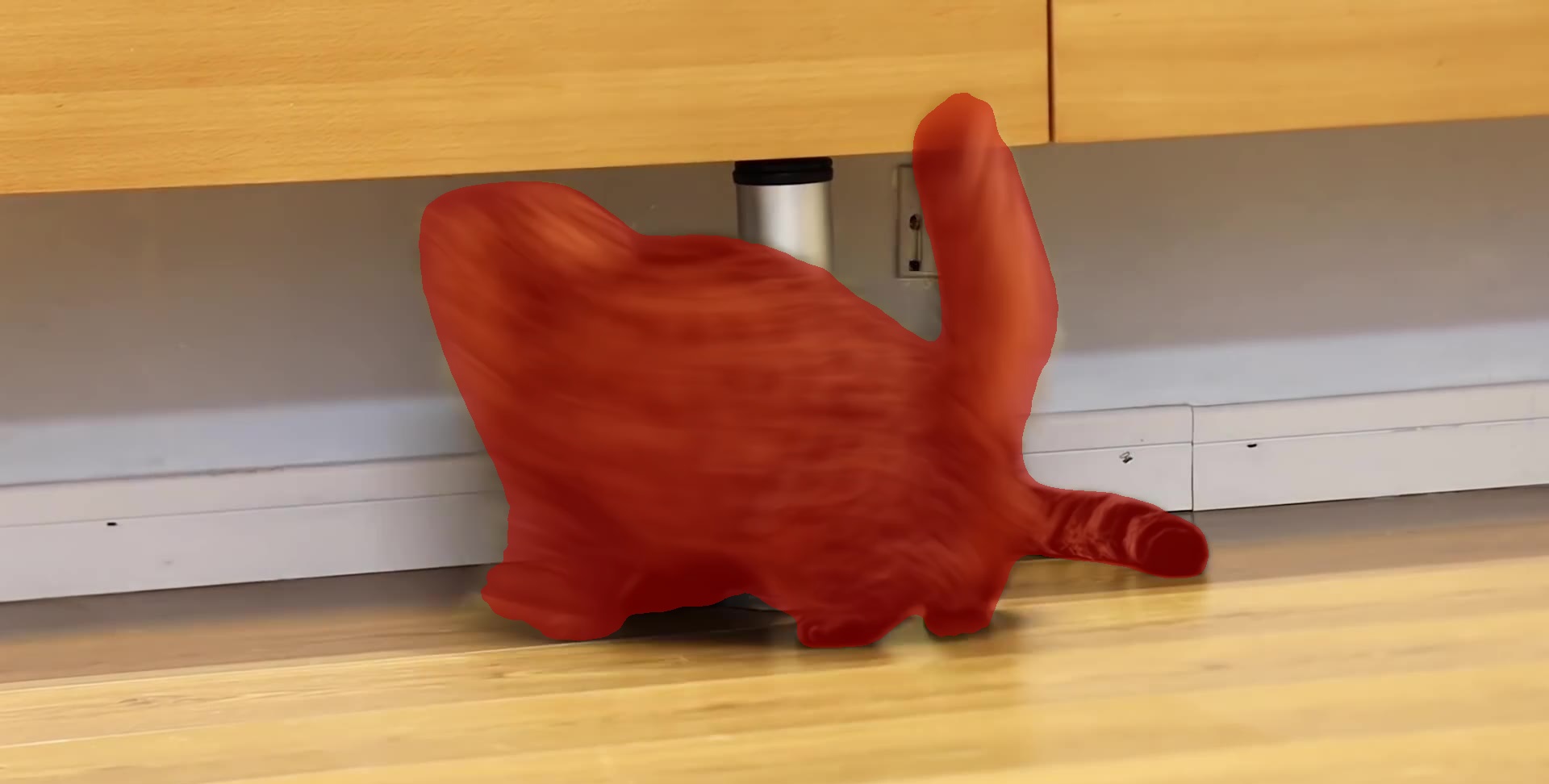} & 
\includegraphics[width=0.12\linewidth]{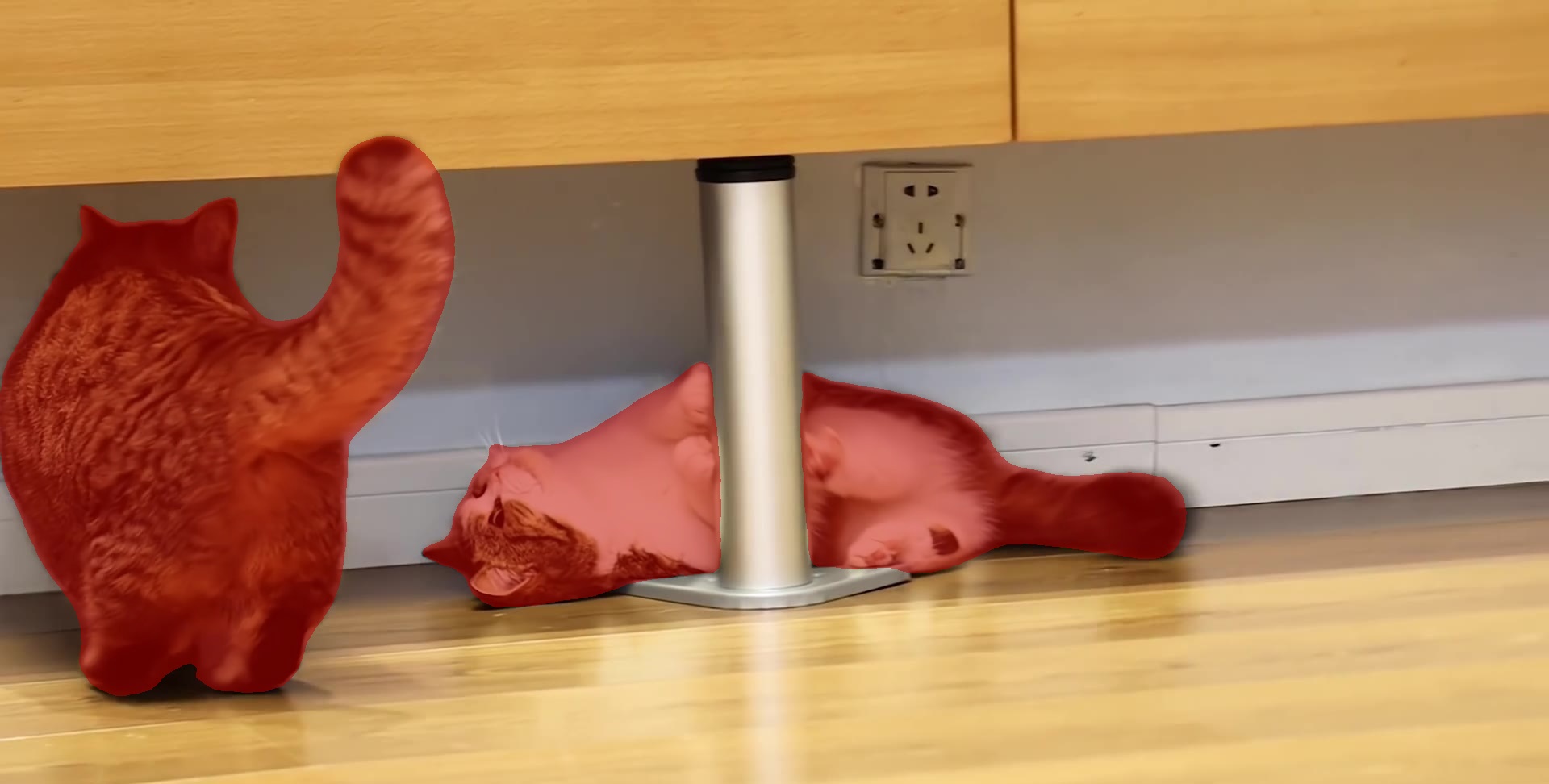} \\[0.5em]

\makebox[0.8cm][c]{\rotatebox{90}{\small \quad Ours}} &
\includegraphics[width=0.12\linewidth]{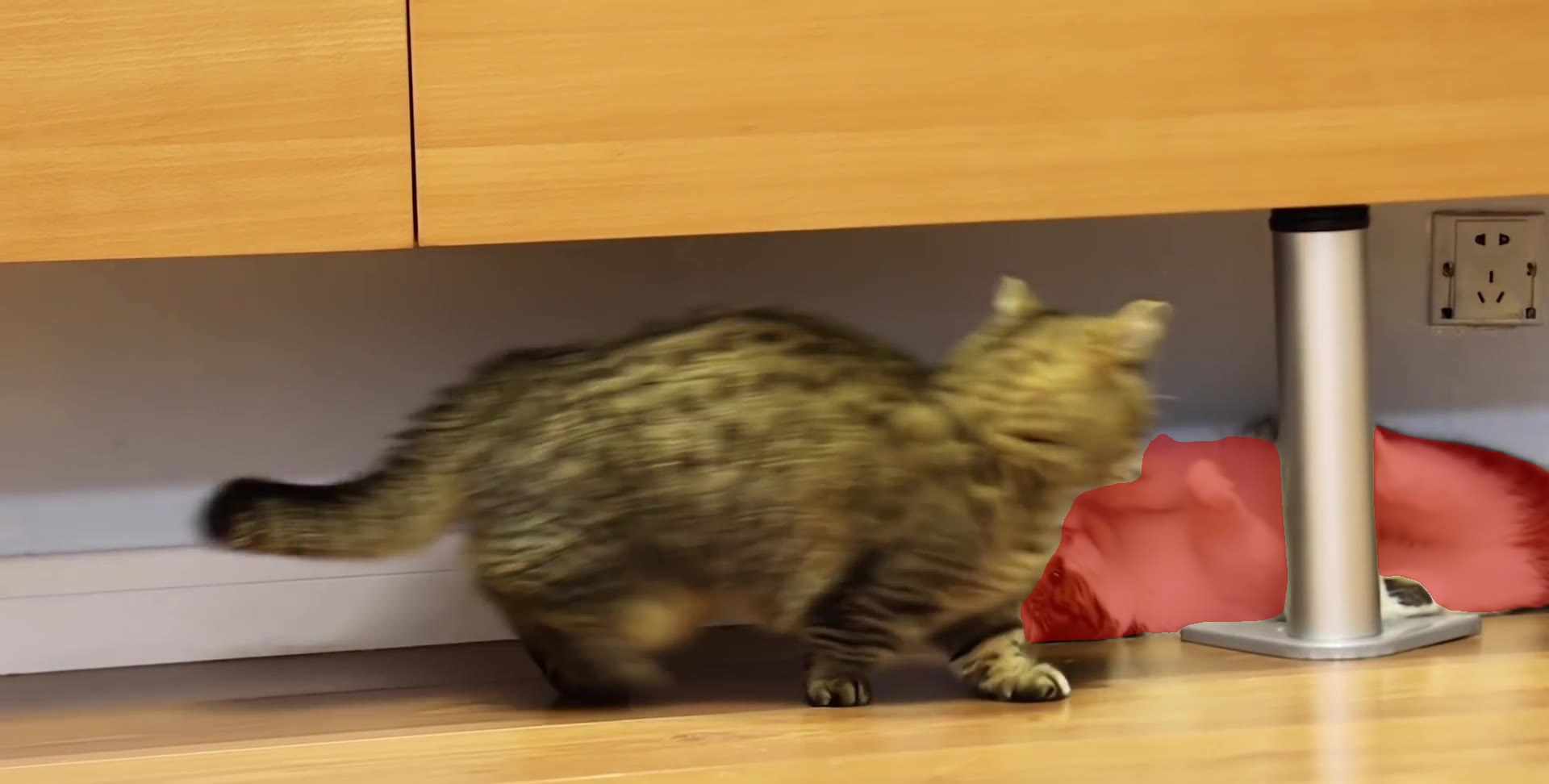} & 
\includegraphics[width=0.12\linewidth]{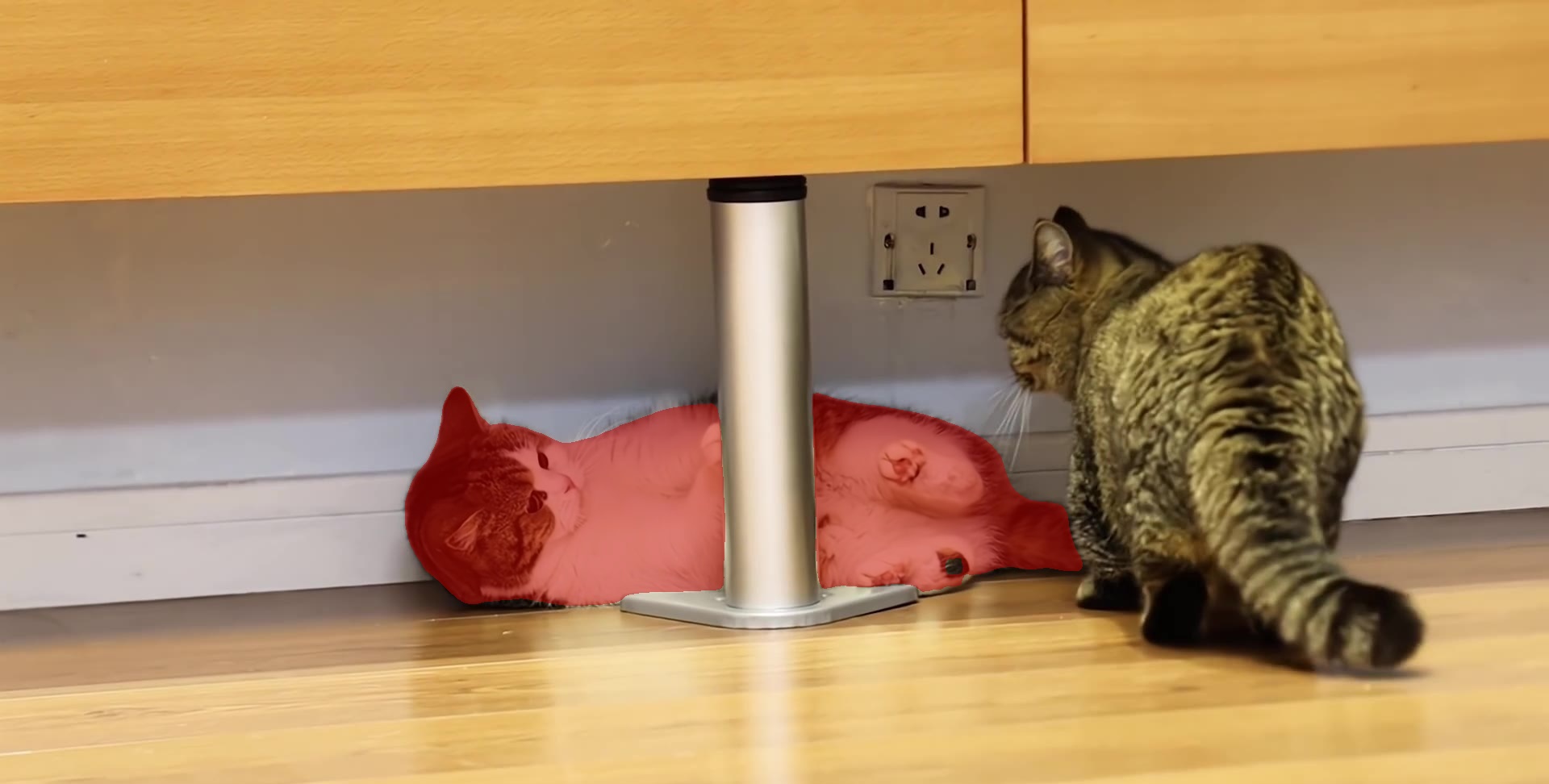} & 
\includegraphics[width=0.12\linewidth]{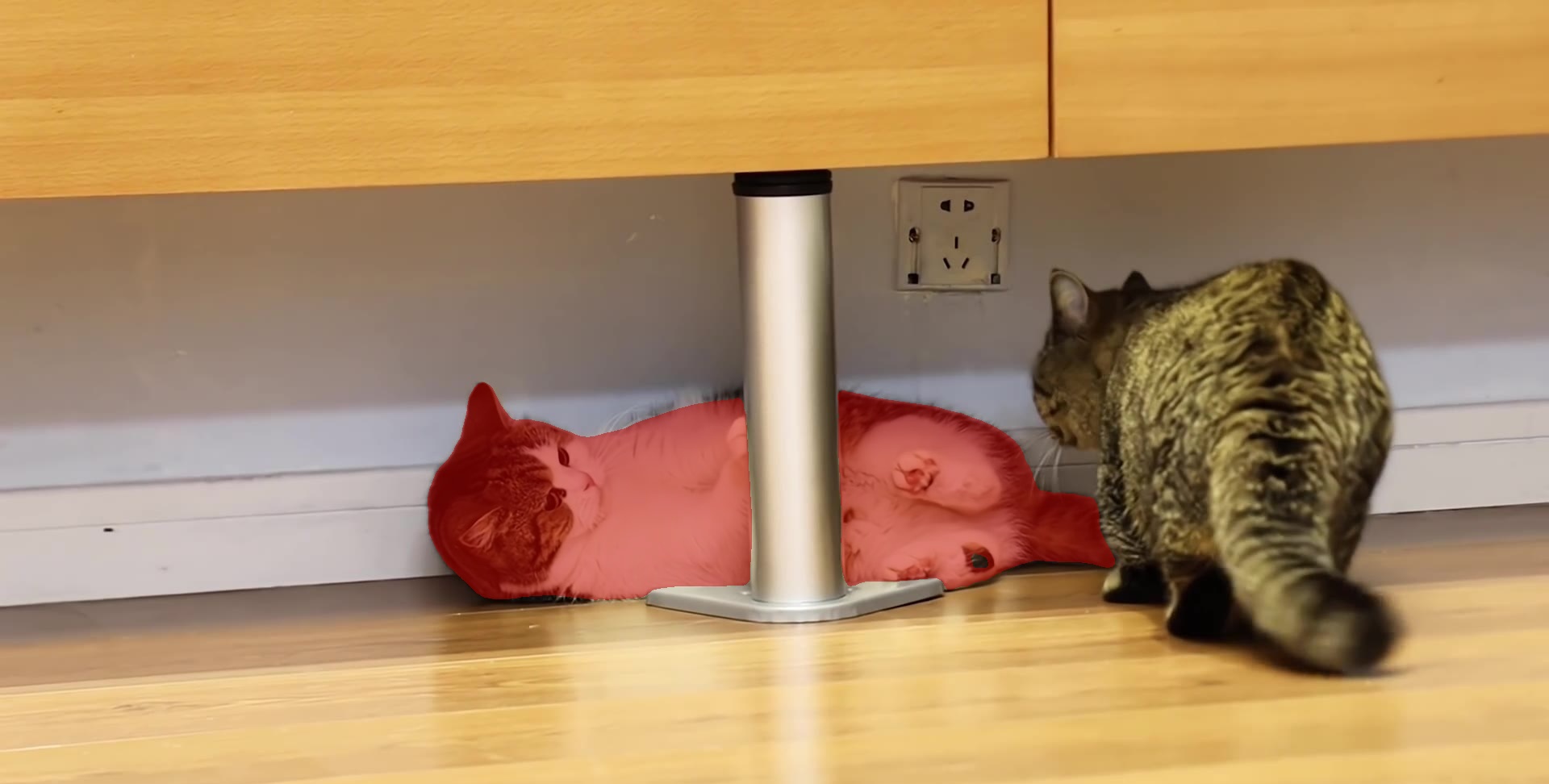} & 
\includegraphics[width=0.12\linewidth]{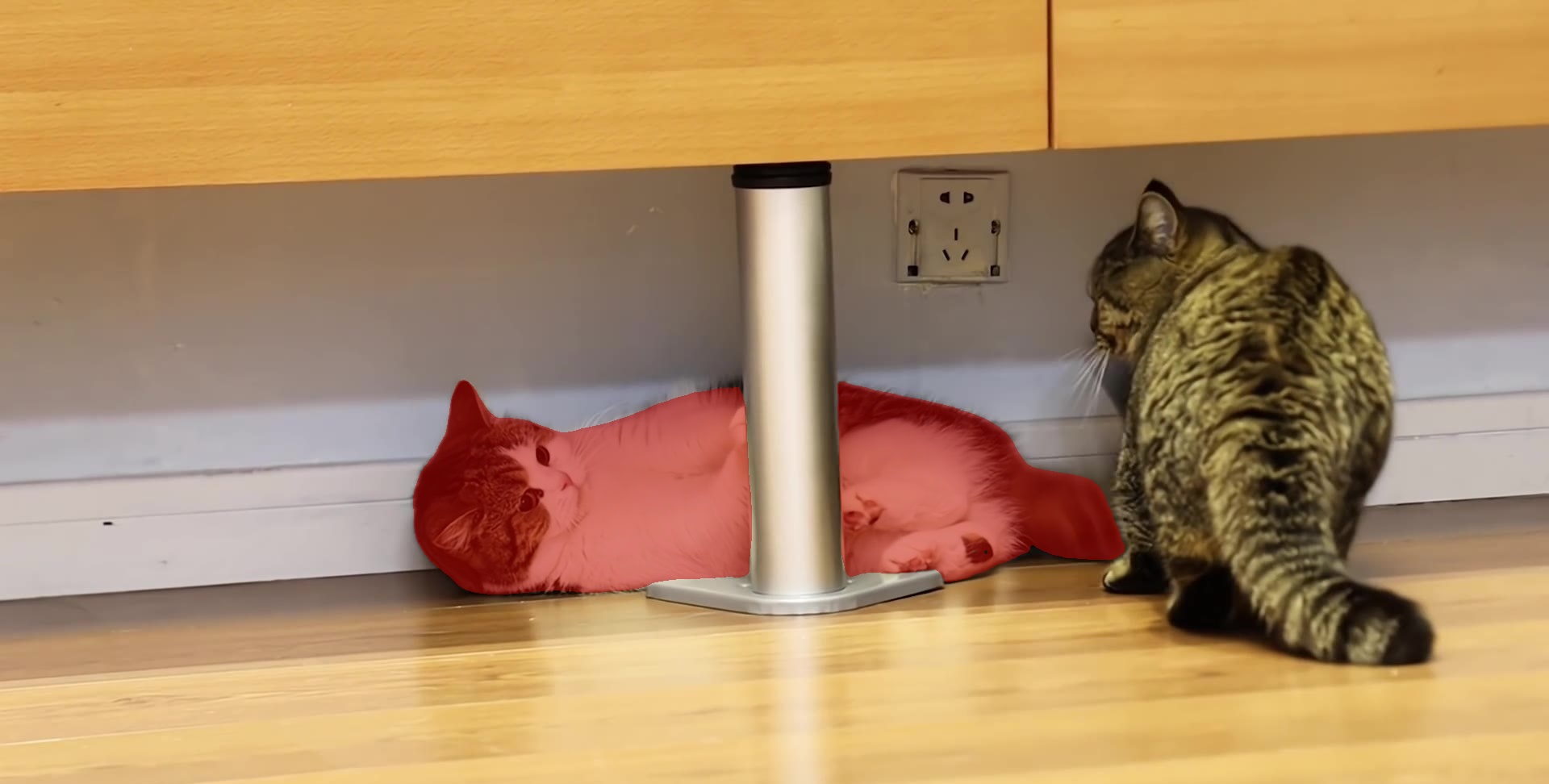} & 
\includegraphics[width=0.12\linewidth]{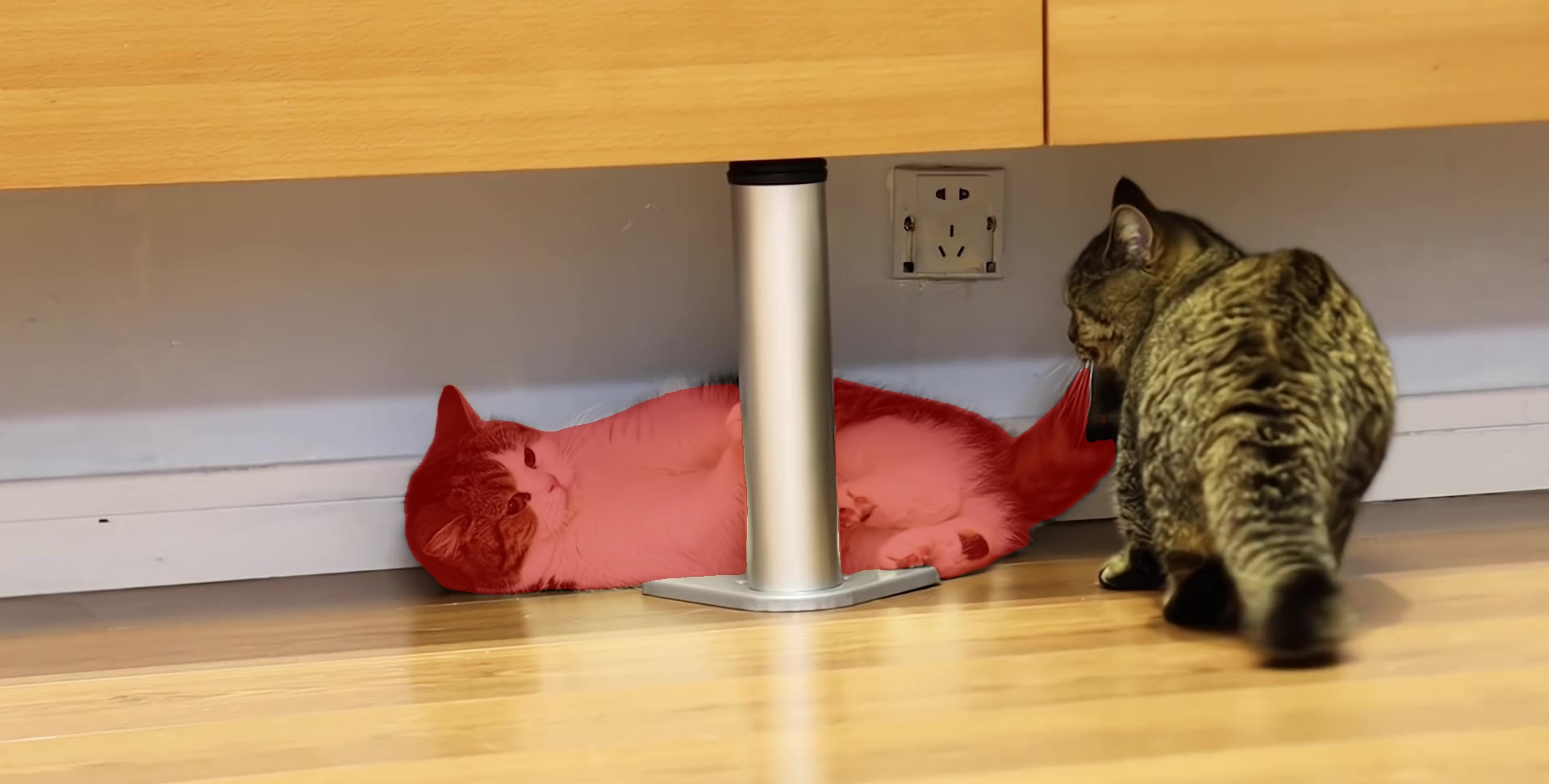} & 
\includegraphics[width=0.12\linewidth]{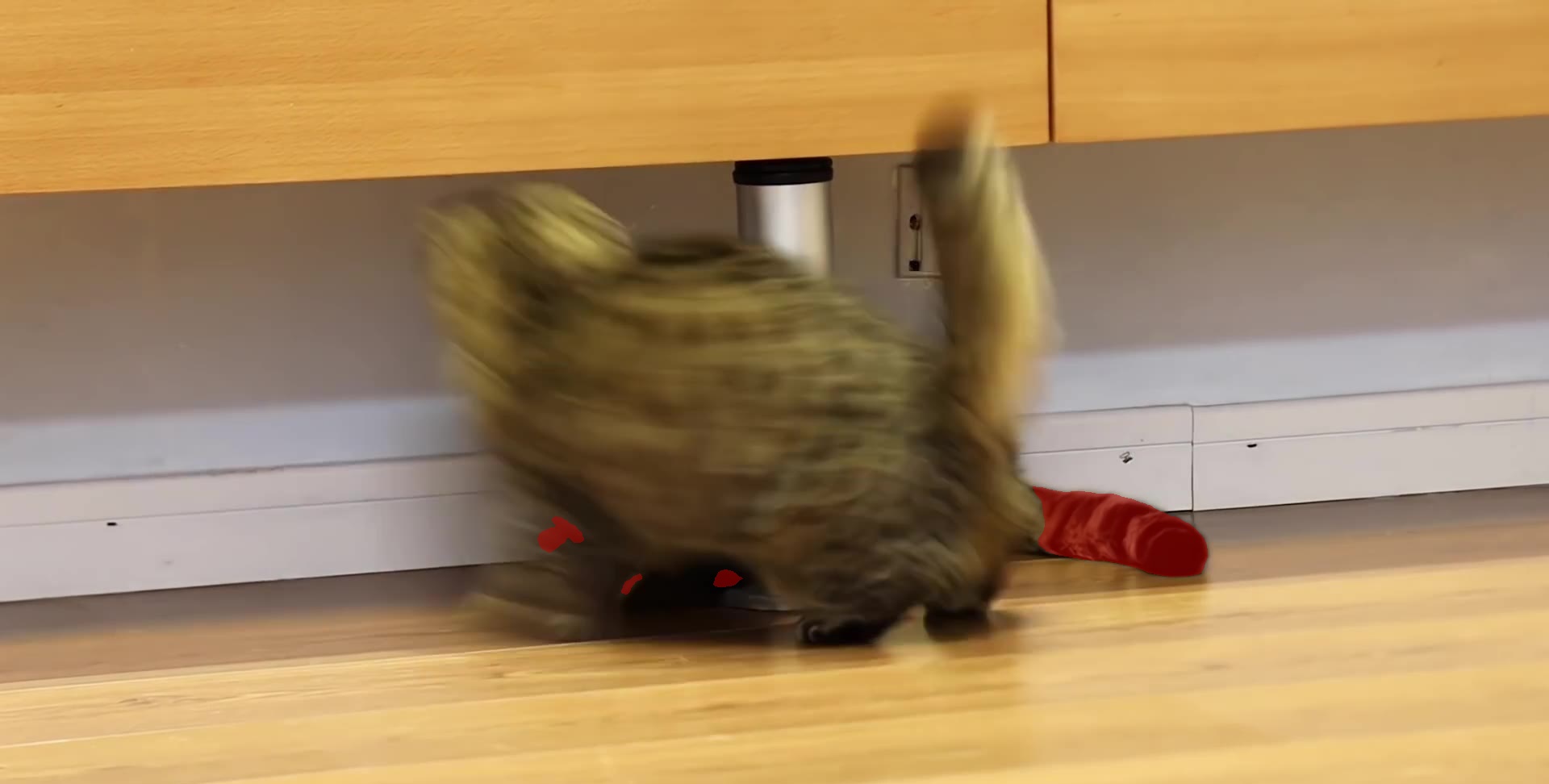} & 
\includegraphics[width=0.12\linewidth]{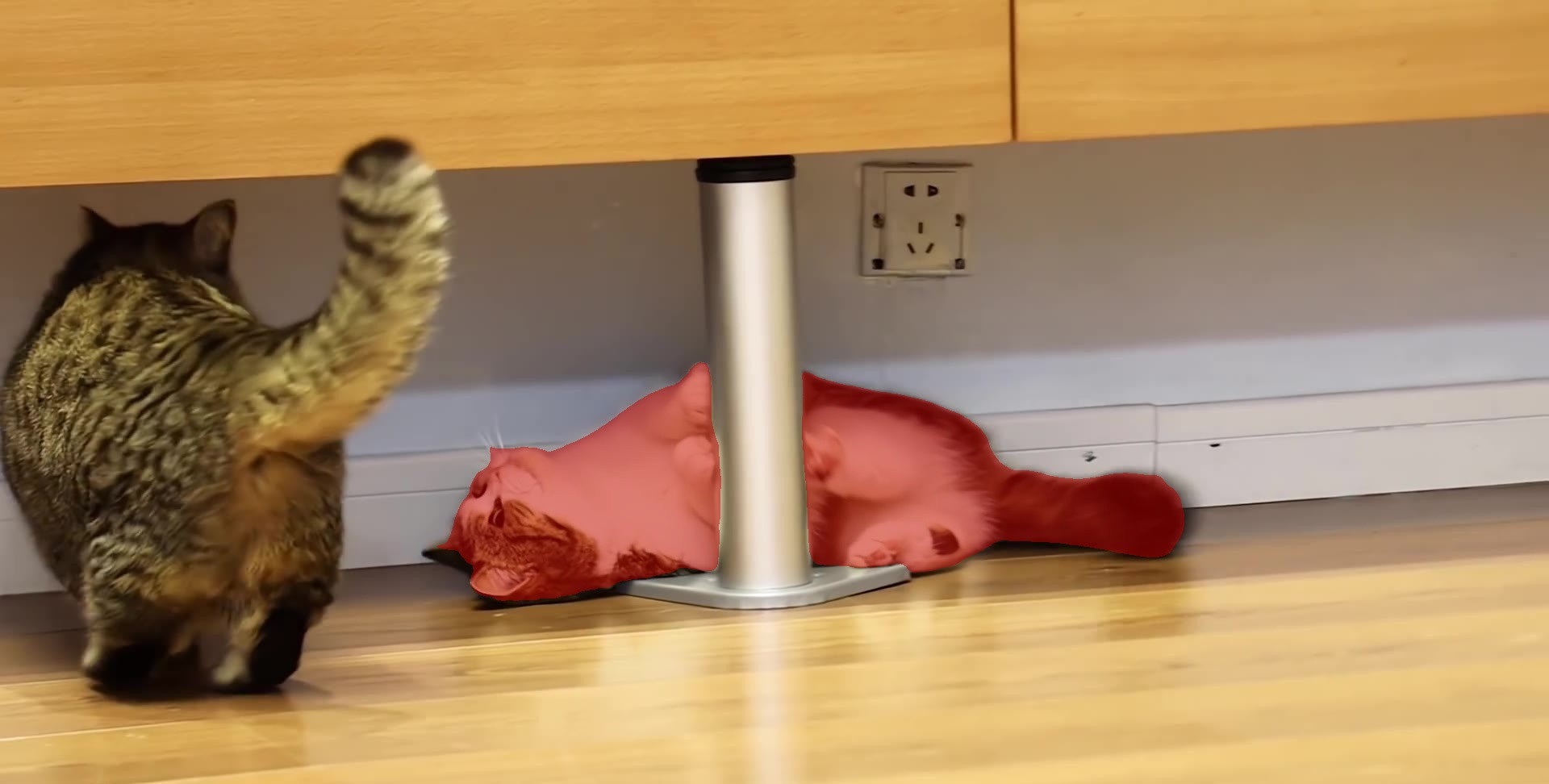} \\
\end{tabular}

\vspace{0.2em}
(a) \textit{"The cat is lounging on the floor, wagging its tail, yet remains stationary."} 

\vspace{1.2em}

\begin{tabular}{c@{\hskip 2pt}c@{\hskip 2pt}c@{\hskip 2pt}c@{\hskip 2pt}c@{\hskip 2pt}c@{\hskip 2pt}c@{\hskip 2pt}c} \makebox[0.8cm][c]{\rotatebox{90}{ \small Baseline\cite{wu2022language}}} &
\includegraphics[width=0.12\linewidth]{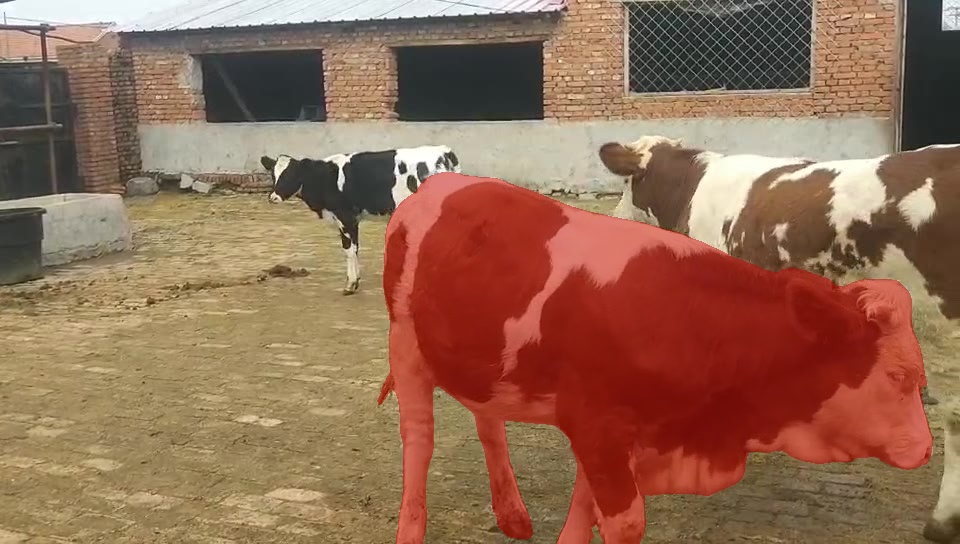} & 
\includegraphics[width=0.12\linewidth]{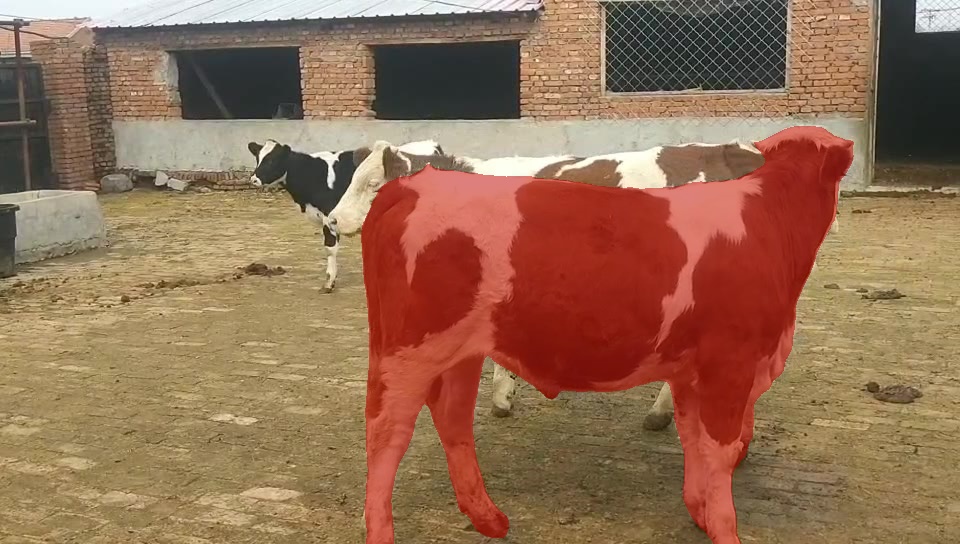} & 
\includegraphics[width=0.12\linewidth]{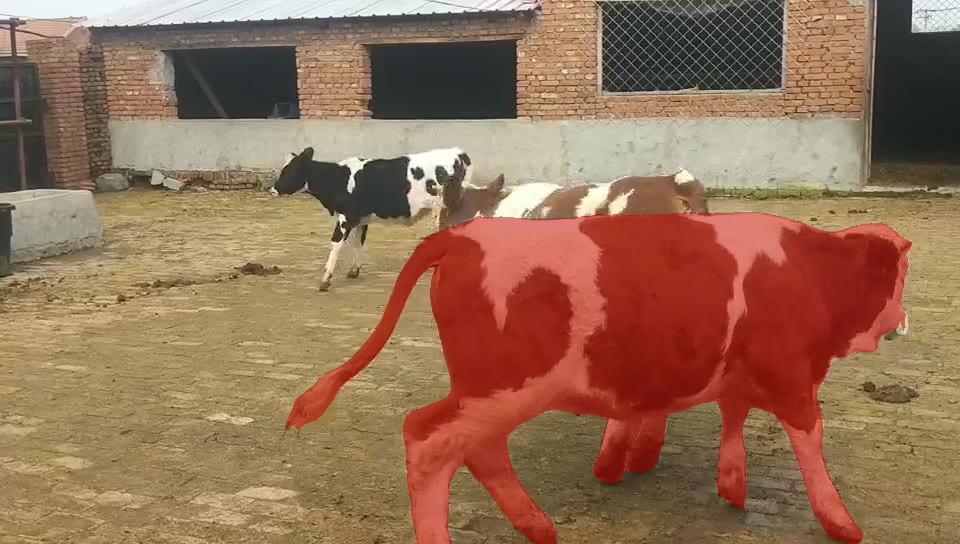} & 
\includegraphics[width=0.12\linewidth]{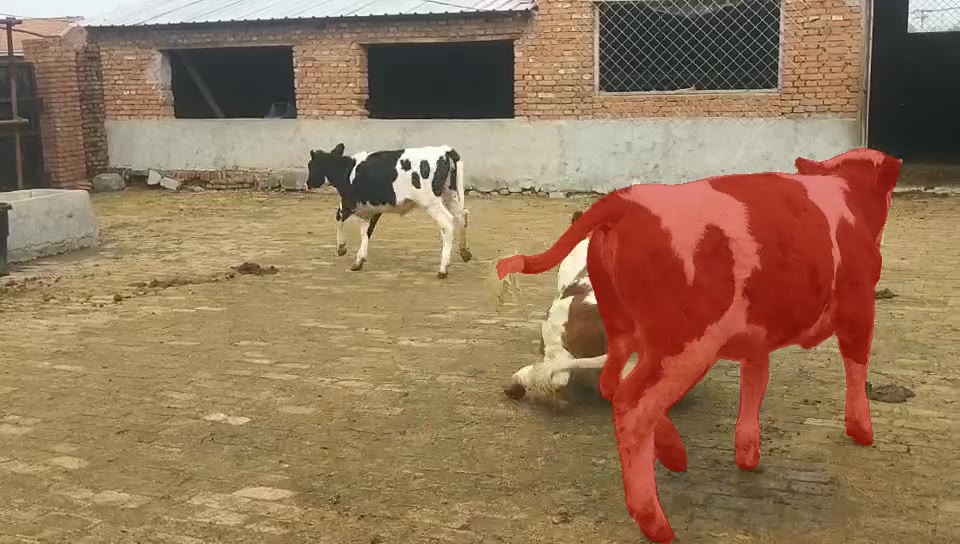} & 
\includegraphics[width=0.12\linewidth]{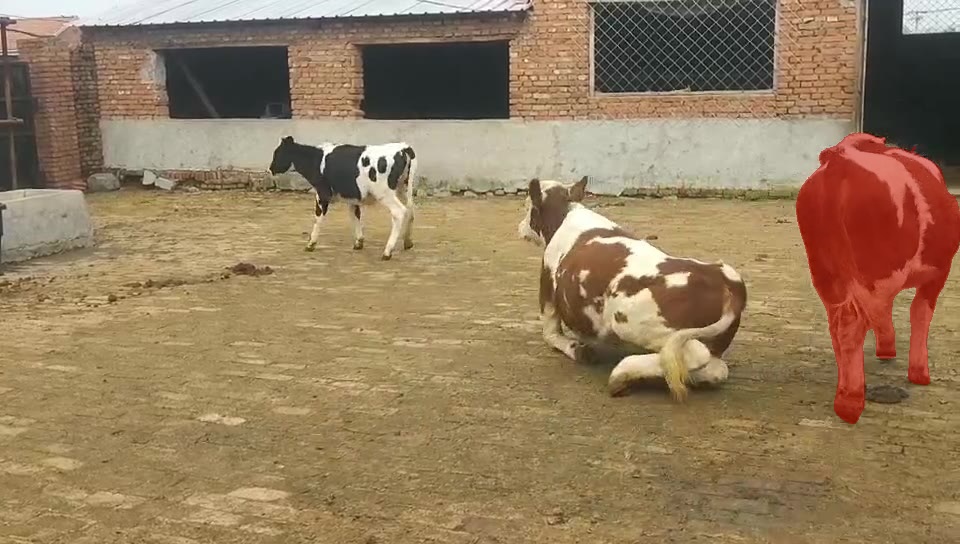} & 
\includegraphics[width=0.12\linewidth]{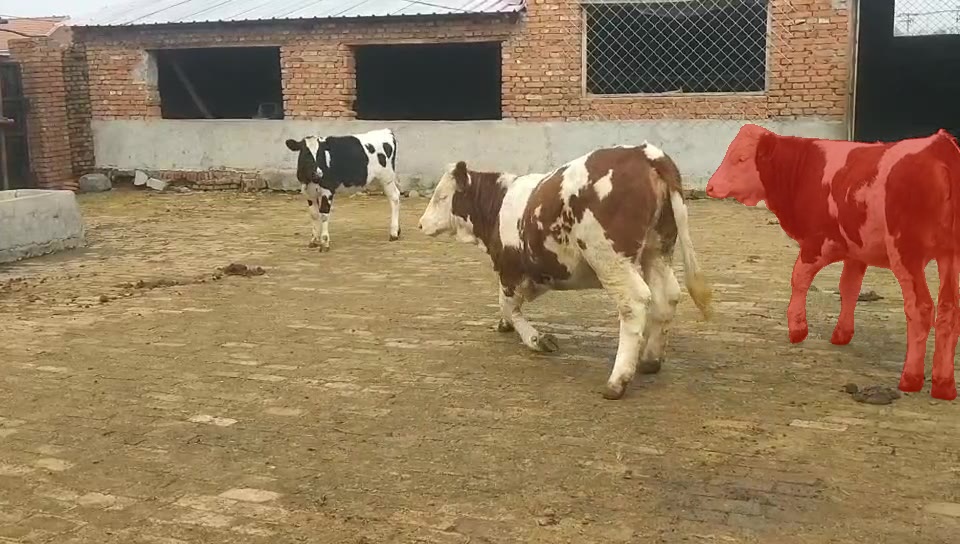} & 
\includegraphics[width=0.12\linewidth]{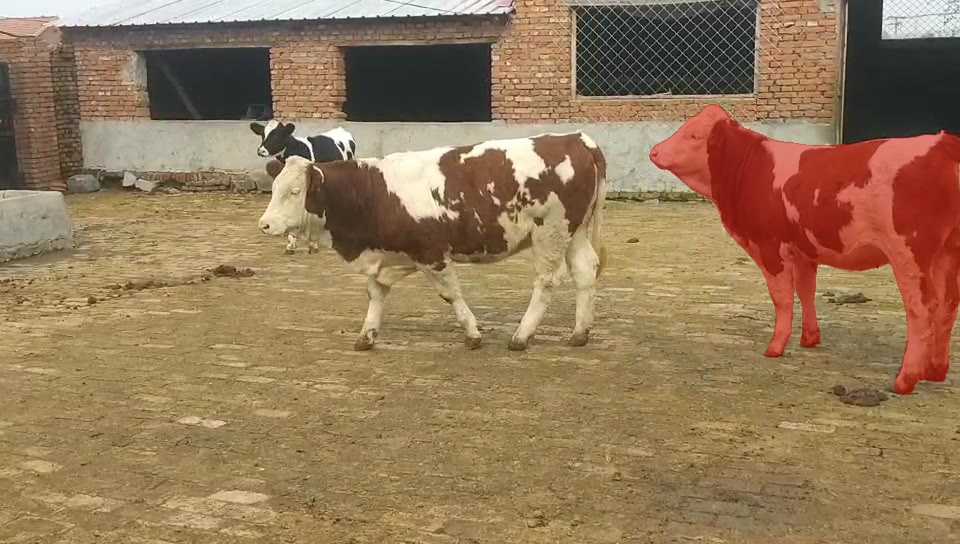}  \\[0.5em]

\makebox[0.5cm][c]{\rotatebox{90}{\small \quad Ours}} &
\includegraphics[width=0.12\linewidth]{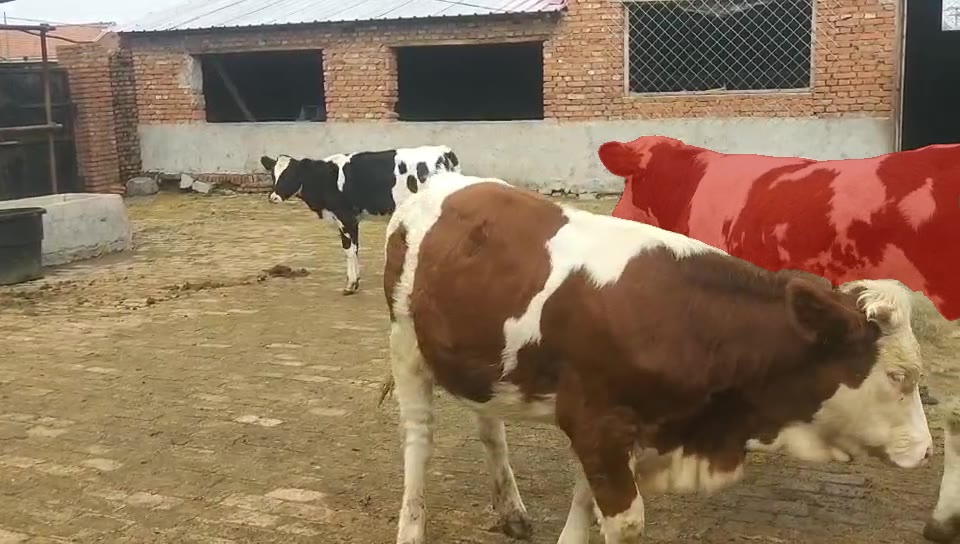} & \includegraphics[width=0.12\linewidth]{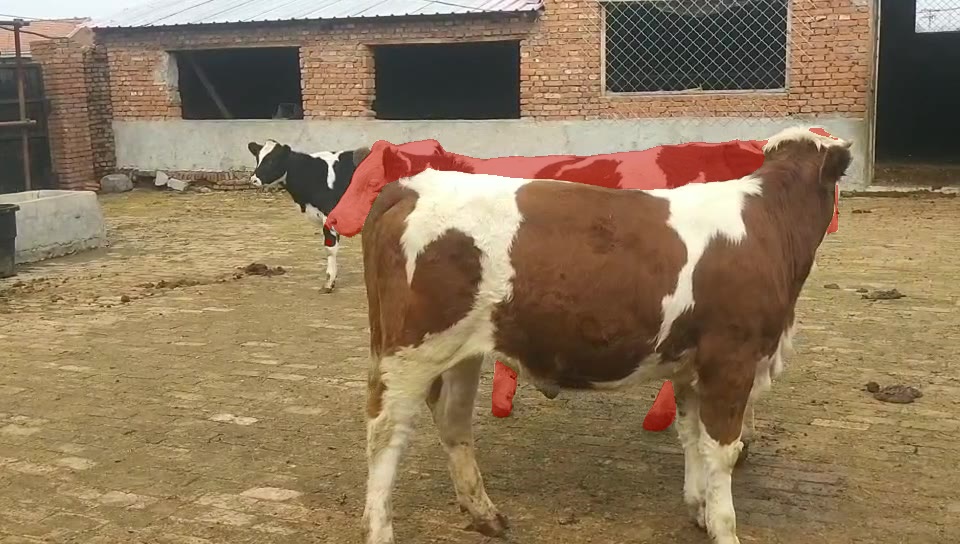} & \includegraphics[width=0.12\linewidth]{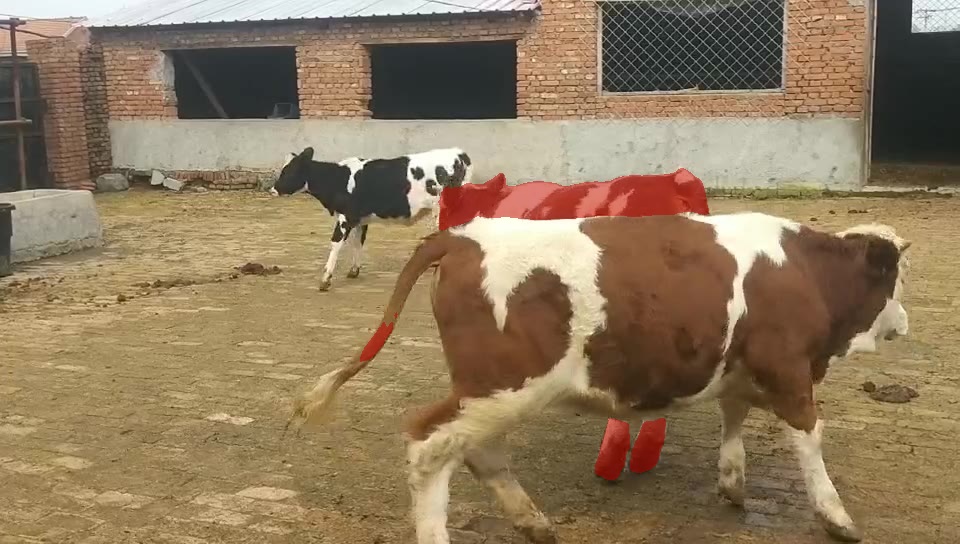} & \includegraphics[width=0.12\linewidth]{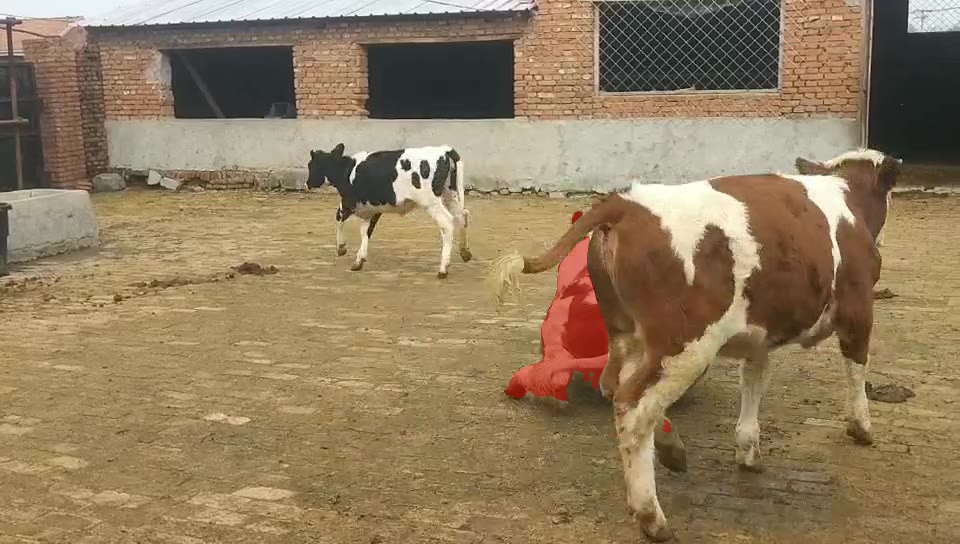} & \includegraphics[width=0.12\linewidth]{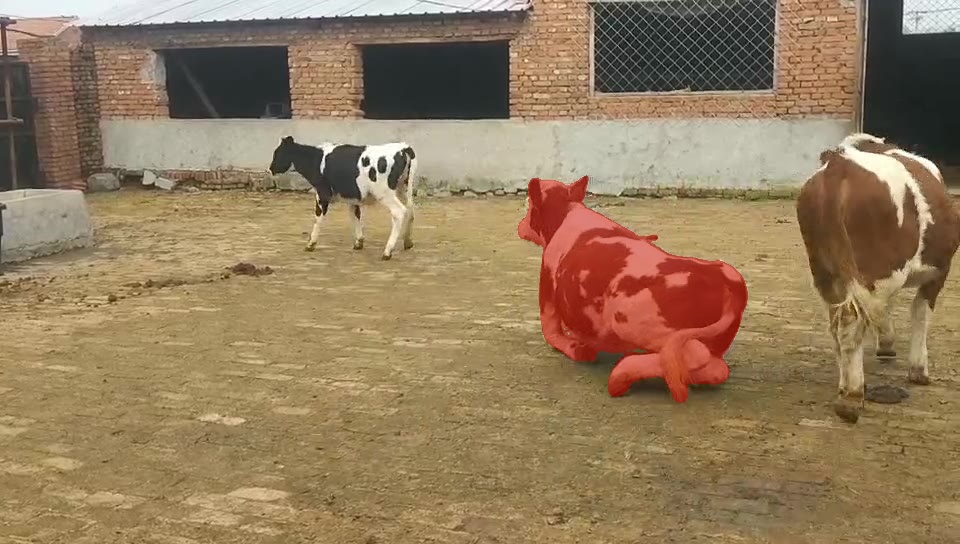} & \includegraphics[width=0.12\linewidth]{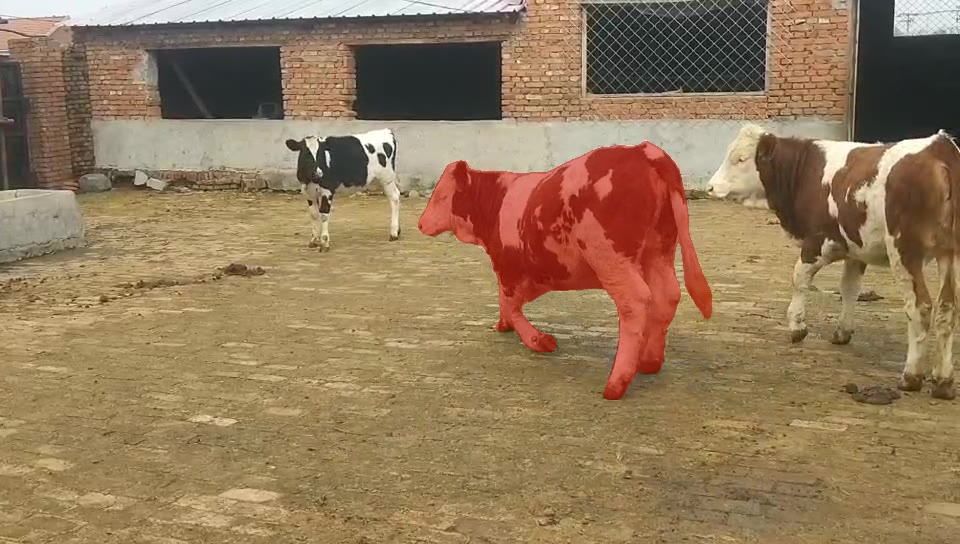} & \includegraphics[width=0.12\linewidth]{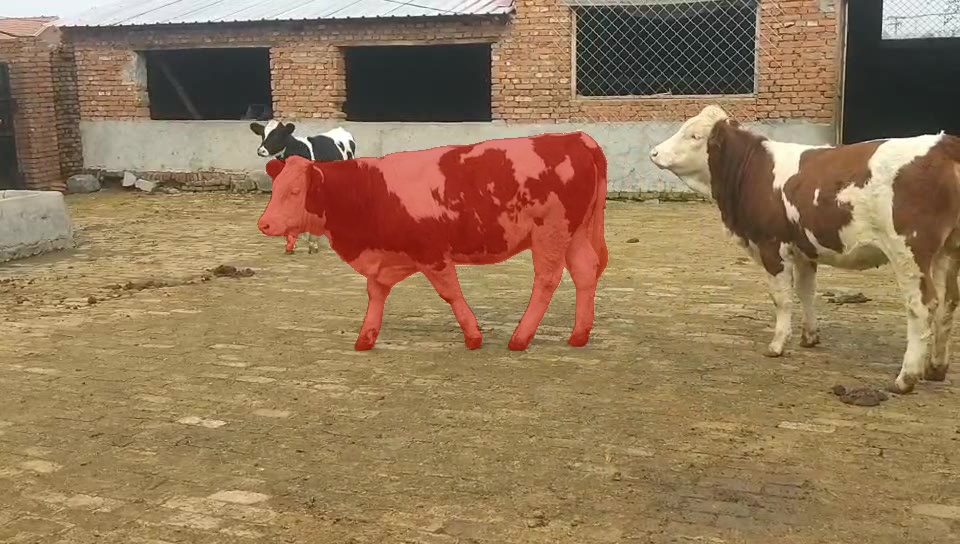}\\
\end{tabular}

\vspace{0.2em}
(b) \textit{"The bull stumbling and hitting the ground."} 

\caption{Qualitative comparisons on the MeViS validation set. }
\label{fig:qual_viz}
\end{figure*}


As summarized in Table \ref{tab1} and \ref{tab2}, we evaluate the proposed TQF on five established R‑VOS benchmarks: Ref‑YouTube‑VOS~\cite{Seo2020}, Ref‑DAVIS17~\cite{khoreva2019video}, A2D‑Sentences and JHMDB‑Sentences~\cite{gavrilyuk2018actor} (extending A2D~\cite{xu2015can} and JHMDB~\cite{jhuang2013towards}), and MeViS~\cite{ding2023mevis}. 
The experiments are carried out using two backbones, \emph{Video‑Swin‑T/Swin‑T} and \emph{Video‑Swin‑B/Swin‑B}, and compared with the most advanced methods. For MeViS, Ref‑YouTube‑VOS, and Ref‑DAVIS17 we report region similarity~($\mathcal{J}$), contour accuracy~($\mathcal{F}$), and their mean~($\mathcal{J}\!\&\!\mathcal{F}$); for A2D‑Sentences and JHMDB‑Sentences we report mean Average Precision (mAP), overall IoU (oIoU), and mean IoU (mIoU).


 \noindent\textbf{A2D-Sentences.}  This dataset emphasizes actor-action interactions with moderate video complexity. Our method achieves the highest performance across all three metrics. With the Video-Swin-T backbone, TQF reaches \textbf{59.2\%} mAP, \textbf{80.4\%} oIoU, and \textbf{71.9\%} mIoU, surpassing the previous best LoSh\cite{yuan2024losh} by \textbf{+1.6\%}, \textbf{+1.1\%}, and \textbf{+0.3\%} points, respectively. With Video-Swin-B, our method further improves to \textbf{61.3\%}/\textbf{82.4\%}/\textbf{73.3\%}, bringing gains of \textbf{+1.4\%}, \textbf{+1.2\%}, and \textbf{+0.2\%} over LoSh.

 \noindent\textbf{JHMDB-Sentences.} On this motion-centric dataset, TQF achieves an mIoU of \textbf{73.2\%} with Video-Swin-T and \textbf{73.7\%} with Video-Swin-B, surpassing DsHmp\cite{he2024decoupling} by \textbf{+1.1\%} and \textbf{+0.8\%}, respectively. We also set new state-of-the-art scores for mAP and oIoU.

\noindent\textbf{Ref-DAVIS17.} This dataset requires accurate multi-object tracking and segmentation with complex backgrounds. TQF achieves a $\mathcal{J}\&\mathcal{F}$ score of \textbf{67.1\%} on Swin-T and \textbf{68.6\%} on Swin-B, outperforming DsHmp\cite{he2024decoupling} by \textbf{+3.1\%} and \textbf{+3.0\%}, respectively.The result validates our  motion aggregation strategy, which builds strong spatial and temporal associations using motion cues.

\noindent\textbf{Ref-YouTube-VOS.} TQF achieves new state-of-the-art results on this large-scale and diverse benchmark. Specifically, we obtain a $\mathcal{J}\&\mathcal{F}$ score of \textbf{65.8\%} (Video-Swin-T) and \textbf{69.1\%} (Swin-B), outperforming the best baseline (LoSh\cite{he2024decoupling}) by \textbf{+2.1\%} and \textbf{+1.9\%}, respectively.


\noindent\textbf{MeVis.} As shown in \textbf{Table~\ref{tab2}}, our TQF\textit{-T} achieves remarkable performance with a $\mathcal{J}\&\mathcal{F}$ score of \textbf{50.1\%}, significantly outperforming previous best DsHmp~\cite{he2024decoupling} by \textbf{+3.7\%}. Notably, the $\mathcal{J}$ and $\mathcal{F}$ scores also see improvements of \textbf{+3.8\%} and \textbf{+3.5\%}, respectively. When scaling up to the TQF\textit{-B} variant, performance further improves to \textbf{52.8\%} in $\mathcal{J}\&\mathcal{F}$, with a region similarity $\mathcal{J}$ of \textbf{49.4\%} and contour accuracy $\mathcal{F}$ of \textbf{56.1\%}.
 Compared to early query-based methods such as URVOS~\cite{Seo2020}, ReferFormer~\cite{wu2022language}, and LMPM~\cite{ding2023mevis}, our method achieves remarkable gains. For instance, TQF\textit{-B} surpasses ReferFormer by \textbf{+21.8\%} and LMPM by \textbf{+15.6\%} in $\mathcal{J}\&\mathcal{F}$, respectively. Even against recent large-scale pretraining methods like Grounded-SAM2~\cite{ren2024groundedsamassemblingopenworld}, TQF\textit{-B} still leads by a margin of \textbf{+12.3\%}.
 These results consistently highlight the effectiveness of our triple-query mechanism in capturing object appearance, intra-frame interactions, and inter-frame motion semantics. The strong performance under both lightweight and large-scale backbones demonstrates the robustness and scalability of TQF in complex referring video segmentation tasks.

\subsection{Qualitative results}

As shown in Figure \ref{fig:qual_viz}, given complex motion-centric referring expressions, our method achieves more accurate and temporally consistent segmentations compared to the baseline (ReferFormer\cite{wu2022language}). In (a), the model precisely captures the static posture and subtle tail motion of the cat. In (b), it robustly follows the bull’s dynamic fall, demonstrating stronger temporal awareness and object interaction understanding.

\subsection{Ablation study}
To evaluate the effectiveness of each component in our framework, we perform a series of ablation experiments on the Ref-YouTube-VOS validation set. All experiments are conducted using Video-Swin-Tiny as the visual backbone, and the model is trained from scratch. Results are summarized in Table~\ref{tab3}, with evaluations grouped into two categories: motion aggregation modules and query design strategies.

\noindent\textbf{Impact of Tokens Aggregation Modules.}   We first analyze the contribution of our proposed Intra-frame Interaction Aggregation (IIA) and Inter-frame Motion Aggregation (IMA) modules. Removing IIA causes a noticeable performance drop of \textbf{7.6\%} in $\mathcal{J}\&\mathcal{F}$, revealing the importance of modeling intra-frame object interactions. Disabling IMA leads to a larger decline of \textbf{8.9\%}, demonstrating the critical role of inter-frame semantic association. When both modules are removed, the model suffers a dramatic decrease of \textbf{16.8\%}, validating that spatial and temporal aggregation are highly complementary. Additionally, removing relative positional encoding (RPE) in the IIA module results in a \textbf{2.1\%} performance drop, indicating the benefit of injecting geometric priors into intra-frame attention.

 \noindent\textbf{Effect of Query Design.}   We further ablate the query initialization strategies. Without cross-modal fusion in the appearance query $N_A$, performance decreases by \textbf{3.5\%}, which shows that alignment with static linguistic attributes is essential. Intra-frame query $N_I$ without sentence-level context fusion results in a mild drop of \textbf{1.3\%}, suggesting global sentence information enhances relational modeling.
 The inter-frame query $N_E$ is particularly sensitive to visual motion priors: removing trajectory fusion degrades performance by \textbf{5.4\%}, underscoring the importance of temporal motion cues. Eliminating positional encoding from $N_E$ also leads to a \textbf{1.6\%} drop, showing that temporal ordering aids cross-frame token consistency.

\section{Conclusion and Future Work}


We introduced Triple Query Former (TQF), which mitigates query selection bias in RVOS by factorising each referring expression into appearance, intra‑frame interaction, and inter‑frame motion queries, combined with motion‑aware aggregation for coherent spatiotemporal reasoning. TQF reaches state‑of‑the‑art performance but falters during long occlusions, where lost appearance and motion cues cause query misalignment and mask drift. Future work will tackle this issue by adding re‑detection and memory‑based persistence to re‑initialize objects after extended occlusions.


\begin{acks}
This work was supported by the Jiangsu Provincial Science and Technology Major Project under Grant BG2024042.
\end{acks}

\bibliographystyle{ACM-Reference-Format}
\bibliography{vos}

\end{document}